\documentclass{article}
\usepackage{iclr2025_conference,times}
\usepackage{array}
\usepackage{hyperref}
\usepackage{subcaption}
\usepackage{url}
\usepackage{multicol}
\usepackage{multirow}
\usepackage{booktabs}
\usepackage{graphicx}
\usepackage{wrapfig}
\usepackage{algorithm}
\usepackage{algpseudocode}
\usepackage{enumitem}
\usepackage{amsmath}
\usepackage{amssymb}
\usepackage{amsthm}
\usepackage{fontenc}
\usepackage{longtable}
\usepackage[margin=1in]{geometry}
\usepackage{xcolor} 
\usepackage{colortbl}
\usepackage{tikz}
\usepackage{adjustbox}
\usetikzlibrary{calc}
\usetikzlibrary{decorations.pathreplacing}
\usetikzlibrary{arrows.meta}
\usetikzlibrary{positioning}
\usetikzlibrary{shapes,fit}

\usepackage{mathtools}    %
\usepackage[most]{tcolorbox}
\usepackage{bm}   
\usepackage{mdframed}  %
\tcbset{
    example/.style={
        colback=gray!10, 
        colframe=gray!40,
        fonttitle=\bfseries
    },
    calculation/.style={
        colback=blue!5, 
        colframe=blue!20,
        fonttitle=\bfseries
    },
    result/.style={
        colback=green!5, 
        colframe=green!20,
        fonttitle=\bfseries
    }
}

\definecolor{codebackground}{rgb}{0.97,0.97,0.97}
\definecolor{codeborder}{rgb}{0.8,0.8,0.8}
\definecolor{codestring}{rgb}{0.639,0.082,0.082}
\definecolor{codekeyword}{rgb}{0.0,0.0,0.7}
\definecolor{codecomment}{rgb}{0.0,0.6,0.0}
\definecolor{placeholdercolor}{rgb}{0.0,0.5,0.5}  %

\lstdefinestyle{maincode}{
    basicstyle=\footnotesize\ttfamily,
    breaklines=true,
    breakatwhitespace=false,
    columns=flexible,
    captionpos=t,
    frame=single,
    framerule=0.5pt,
    rulecolor=\color{codeborder},
    backgroundcolor=\color{codebackground},
    numbers=left,
    numberstyle=\tiny\color{gray},
    numbersep=5pt,
    showstringspaces=false,
    tabsize=2,
    keywordstyle=\color{codekeyword}\bfseries,
    stringstyle=\color{codestring},
    commentstyle=\color{codecomment}\itshape,
    xleftmargin=17pt,
    framexleftmargin=17pt,
    belowskip=0.5\baselineskip,
    aboveskip=0.5\baselineskip
}

\lstdefinestyle{compactcode}{
    basicstyle=\scriptsize\ttfamily,
    breaklines=true,
    breakatwhitespace=false,
    columns=flexible,
    captionpos=t,
    frame=single,
    framerule=0.5pt,
    rulecolor=\color{codeborder},
    backgroundcolor=\color{codebackground},
    numbers=left,
    numberstyle=\tiny\color{gray},
    numbersep=5pt,
    showstringspaces=false,
    tabsize=2,
    keywordstyle=\color{codekeyword},
    stringstyle=\color{codestring},
    commentstyle=\color{codecomment}\itshape,
    xleftmargin=17pt,
    framexleftmargin=17pt,
    belowskip=0.5\baselineskip,
    aboveskip=0.5\baselineskip
}

\lstdefinestyle{templatecode}{
    basicstyle=\footnotesize\ttfamily,
    breaklines=true,
    breakatwhitespace=false,
    columns=flexible,
    captionpos=t,
    frame=single,
    framerule=0.5pt,
    rulecolor=\color{codeborder},
    backgroundcolor=\color{codebackground},
    numbers=left,
    numberstyle=\tiny\color{gray},
    numbersep=5pt,
    showstringspaces=false,
    tabsize=2,
    keywordstyle=\color{codekeyword}\bfseries,
    stringstyle=\color{codestring},
    commentstyle=\color{codecomment}\itshape,
    moredelim=[is][\color{placeholdercolor}\bfseries]{<<}{>>},  %
    xleftmargin=17pt,
    framexleftmargin=17pt,
    belowskip=0.5\baselineskip,
    aboveskip=0.5\baselineskip
}

\newmdenv[
    linewidth=0.5pt,
    roundcorner=3pt,
    backgroundcolor=codebackground,
    innerleftmargin=5pt,
    innerrightmargin=5pt,
    innertopmargin=5pt,
    innerbottommargin=5pt
]{codehighlight}

\usetikzlibrary{arrows.meta,calc,decorations.pathreplacing,positioning,shapes,fit,backgrounds,shadows,trees,mindmap}

\title{\logo \ema: A Self-Optimizing System for Seamless LLM Selection and Integration}

\author{
  Soham Shah, \quad Kumar Shridhar, \quad Surojit Chatterjee, \quad Souvik Sen \\
  \includegraphics[height=1em]{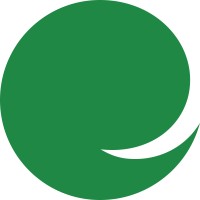}~Ema Unlimited, Inc. \\
  \texttt{\{soham,shridhar,surojit,souvik\}@ema.co}
}

\newcommand{\ema}{\texttt{EMAFusion\textsuperscript{\tiny TM}}}
\newcommand{\logo}{\includegraphics[height=0.9cm, trim=0 65 0 0, clip]{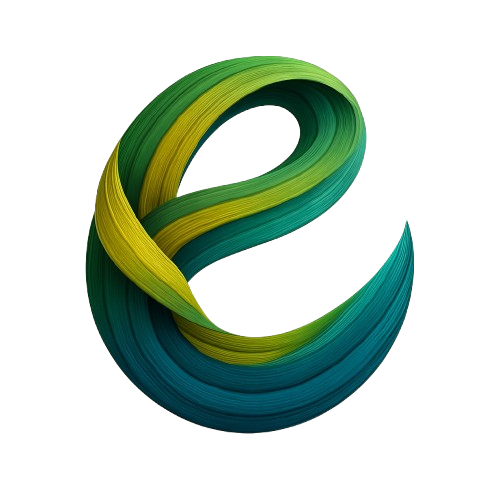}}
\iclrfinalcopy %
\begin{document}

\title{\logo \ema: A Self-Optimizing System for Seamless LLM Selection and Integration}

\maketitle

\begin{abstract}
While recent advances in large language models (LLMs) have significantly enhanced performance across diverse natural language tasks, the high computational and financial costs associated with their deployment remain substantial barriers. Existing \emph{routing} strategies partially alleviate this challenge by assigning queries to cheaper or specialized models, but they frequently rely on extensive labeled data or fragile task-specific heuristics. Conversely, \emph{fusion} techniques aggregate multiple LLM outputs to boost accuracy and robustness, yet they often exacerbate cost and may reinforce shared biases.

We introduce \ema, a new framework that self-optimizes for seamless LLM selection and reliable execution for a given query. Specifically, \ema\ integrates a \emph{taxonomy-based} router for familiar query types, a \emph{learned} router for ambiguous inputs, and a \emph{cascading} approach that progressively escalates from cheaper to more expensive models based on multi-judge confidence evaluations. 
Through extensive evaluations, we find \ema\ outperforms the best individual models by over 2.6 percentage points (94.3\% vs. 91.7\%), while being 4X cheaper than the average cost. \ema\ further achieves a remarkable 17.1 percentage point improvement over models like GPT-4 at less than 1/20th the cost. Our combined routing approach delivers 94.3\% accuracy compared to taxonomy-based (88.1\%) and learned model predictor-based (91.7\%) methods alone, demonstrating the effectiveness of our unified strategy. Finally, \ema\ supports flexible cost-accuracy trade-offs, allowing users to balance their budgetary constraints and performance needs.
\end{abstract}

\section{Introduction}

Large language models (LLMs) have achieved transformative performance improvements across various natural language processing tasks, including question-answering, translation, and reasoning \cite{achiam2023gpt, team2024gemini, anthropic2024claude3, grattafiori2024llama}. However, the substantial computational and financial burdens associated with deploying these models remain significant barriers to widespread practical adoption. As a result, two prominent solution strategies have evolved to address cost and performance trade-offs: \emph{LLM routing} and \emph{LLM fusion}.

\paragraph{Limitations of Existing Routing}  
Routing methods reduce overhead by intelligently selecting a model based on each query’s complexity. For instance, Eagle Router~\cite{zhao2024eagle} uses heuristics to rank models with Elo-like ratings, while IBM’s router~\cite{shnitzer2023large} and GraphRouter~\cite{feng2024graphrouter} rely on learned classifiers or graph-based models to choose the best-performing LLM per query. However, most existing routing methods depend heavily on either extensive labeled training data or on rigid heuristics that risk failing on ambiguous or out-of-distribution (OOD) inputs. Additionally, focusing purely on routing may neglect performance gains achievable by integrating multiple model outputs for challenging problems.

\paragraph{Limitations of Existing Fusion.}
Fusion approaches, by contrast, aim to elevate reliability by merging predictions from multiple LLMs. Voting- or ensemble-based methods such as Self-Consistency~\cite{wang2022self} or LLM-Debate~\cite{du2023improving} have achieved notable accuracy boosts. Yet, these techniques often ignore cost constraints (by calling many models or running repeated generations) and can inadvertently reinforce shared biases among models if the aggregation mechanism is not robust. Consequently, while fusion offers potential accuracy gains, its high cost and sensitivity to model agreement remain practical drawbacks.

\begin{figure}
    \centering
    \includegraphics[width=0.99\linewidth]{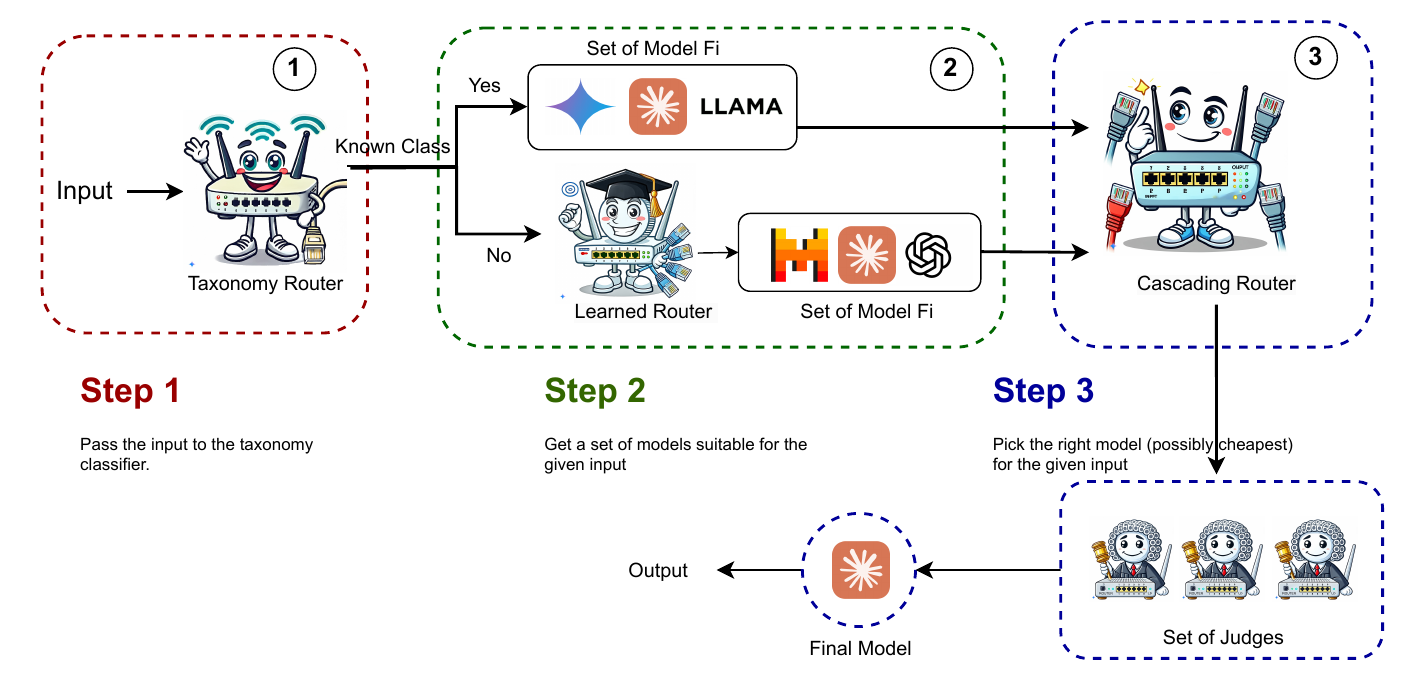}
    \caption{Overview of the \ema\ pipeline. An incoming query first goes to a \emph{Taxonomy Router} to check if it belongs to a known class. If so, it is routed directly to a set of suitable models; otherwise, a \emph{Learned Router} selects a candidate set. Finally, a \emph{Cascading Router} picks from the candidate models in order of cost/performance, and a series of judges verifies the output.}
    \label{fig:main}
\end{figure}

Building on the strengths and addressing the shortcomings of these paradigms, we introduce \ema, a hybrid methodology that \emph{combines intelligent routing with a cost-aware selection strategy}. As illustrated in Figure~\ref{fig:main}, \ema\ starts by \emph{decomposing} a task into subproblems and employs a \emph{taxonomy-based router} to handle queries that fall within known categories and adhere to human preference. If the query is ambiguous or OOD, a \emph{learned router} is used to score the suitability of each candidate model. Finally, \ema\ adopts a \emph{cascading} approach to balance cost and accuracy, escalating from cheaper to more expensive models with \emph{novel judging criteria}. 
This proposed judging-based fusion mitigates biases (e.g., the pitfalls of simple majority voting) by utilizing multiple independent judges to assign aggregated confidence scores depending on the task type. 

Through this presented framework, \ema\ achieves significantly better performance and cost efficiency than either routing or fusion-only baselines. In our empirical evaluations over various tasks ranging from instruction following to reasoning and code evaluations, \ema\ outperforms the best individual model by 2.6 percentage points (94.3\% vs. 91.7\% for O3 Mini), while providing a 10.6\% improvement over the average LLM, at less than $1/3^{rd}$ the cost (\$5.21 vs. \$16.29 per 1000 prompt samples). This hybrid approach demonstrates clear advantages over single-strategy methods—our combined taxonomy and learned router with cascading achieves 94.3\% accuracy, compared to 88.1\% for taxonomy-based routing alone and 91.7\% for predictor-based methods. These gains stem from the complementary nature of our components: taxonomy-based routing provides structured domain knowledge for well-understood queries, the learned router adapts to novel patterns, and the cascading mechanism ensures optimal cost-performance balance. The remainder of this paper details our methodology, presents our experimental results across diverse tasks, and analyzes the specific contributions of each component to \ema's overall performance.

\section{Related Work}
\subsection{LLM Routing}
Using a single large model for all queries is expensive. Routing frameworks seek to optimize the \textit{accuracy-vs.-cost} trade-off by calling small or specialized models for simpler tasks and large models for complex queries \cite{ong2024routellm}. This can reduce both inference time and API costs without sacrificing performance.

\paragraph{Taxonomy based Routing} The simplest way to solve the issue is by using a \textit{Hard-coded rules or some heuristics} to detect certain query types (e.g., code vs.\ casual chat) and select a corresponding specialized model. However, this will require a strong domain knowledge of each task and may fail on ambiguous queries. Eagle Router \cite{zhao2024eagle} uses such a heuristic and ranks LLMs by skill using an Elo rating system, and can select models quickly without training overhead. \citet{Yang-medical-routing} propose LLM-Synergy for medical QA, which uses cluster-based dynamic model selection, essentially grouping questions by context and choosing the most suitable LLM’s answer for each query. Similarly, we propose a taxonomy-based routing as the first step in the routing process to choose the appropriate models based on the group a query belongs to, or if the user has specified some preferred models.

\paragraph{Learned Model Routing} An alternative is to train a \textit{Learned Router Model}, which can be a supervised classifier or regression model to predict the best model for each input. This will require labeled data from multiple tasks to train this classifier. 
IBM's router \cite{shnitzer2023large} uses large-scale NLP benchmarks to train a classifier that picks the highest-performing model per query. Similarly, GraphRouter \cite{feng2024graphrouter} constructs a heterogeneous graph of tasks, queries, and LLMs to capture rich relationships; a graph-based model then predicts which LLM will give the best trade-off of effect (quality) and cost for a query. \citet{aljundi} introduced Expert Gate, a lifelong learning model that adds a new expert network for each task and trains gating autoencoders to decide which expert to use at test time. 

\paragraph{Cascading} While accuracy is one of the most important parameters to optimize for routing, it might be expensive to choose the most accurate models. A common cost-saving strategy is to cascade from cheaper models to expensive ones only when needed \citep{chen2023frugalgpt, chen2024model}. While past works either use self-evaluation \cite{chen2024model} or a smaller model as a judge \citep{chen2023frugalgpt, shridhar-etal-2024-art}, a combination of prompt adaptation, approximator models, and cascaded fallback to powerful LLMs yields better cost–performance tradeoffs. While there have been attempts to combine routing and cascading \cite{dekoninck2024unified}, the approach relies on a training dataset to optimize the hyperparameters associated with cascade routing. On the other hand, our proposed cascading approach is judged by a series of LLMs as independent judges providing confidence scores for different metrics, and the final score is an aggregator over all metrics. This prevents the self-bias aspect from self-judging and removes the complexity of training any model. Finally, cascading has some similarities with speculative decoding, where a smaller model generates the draft of the answer and a larger model corrects or finishes it \cite{leviathan2023fast}. Our cascading approach allows the smaller model to generate the full answer, and then it is judged by a series of expert models across various metrics. 

\paragraph{Other approaches for Routing}
An alternative line of work embeds the routing mechanism within the model’s architecture. Mixture-of-Experts (MoE) models (e.g., Switch Transformer \cite{fedus2021switch}, GLaM \cite{du2022glam}) consist of many expert sub-networks, with a learned gating network that routes each input (or even each token) to one or a few expert networks. This intra-model routing achieves the effect of a huge model (many parameters across experts) while each input only activates a subset, keeping computation manageable. While MoE approaches are implemented at the neural layer level, they exemplify the principle of routing to reduce cost: only use heavy computation for those inputs that need it. Our LLM routing systems can be seen as an extension of this idea to multiple distinct models, not just experts in one network. 

However, LLM routing does not work in all cases, and \citet{srivatsa2024harnessing} found that their trained routing model, despite the theoretical potential to beat all individual LLMs, in practice only matched the top single model’s performance for reasoning tasks. Finally, an important aspect that \citet{varangot2025doing} emphasizes is the complementarity of the model pool: having models with diverse strengths (different sizes, domains, training data) is key to unlocking significant performance gains through routing. If all models are similar, routing doesn’t help much; the biggest win comes when at least one option can handle certain queries much more cheaply or accurately than the others.

\subsection{LLM Fusion} 
Whereas routing chooses one model per query, LLM fusion techniques seek to combine multiple models’ outputs for a single query. The intuition is that each model may contribute useful information or perspectives, so merging their answers could yield a more accurate or comprehensive result than any single model alone. Fusion methods often aim to maximize performance (sometimes at the expense of additional compute cost), making them popular in settings where quality is paramount. Moreover, LLM Fusion can also enhance robustness: if one model hallucinates or errs, others might correct it, and a fusion mechanism can down-weight outliers. In tasks requiring high reliability (e.g., medical QA), leveraging multiple models can provide an extra layer of validation or confidence. 

\paragraph{Voting and Ensemble}
A straightforward fusion approach is to use ensembles of LLMs or prompts and then apply voting or ranking to select the best answer. This can be as simple as majority voting: pose the query to several models (or run one model multiple times with varied prompts or random seeds) and see which answer is most common. One such popular approach is \emph{Self-Consistency} \cite{wang2022self}, where multiple paths are generated from an LLM, and the answer is selected that occurs most consistently among those paths. \citet{li2024more} further confirmed that performance scales with the number of independent agents sampled – essentially, the more independent attempts the model makes, the higher the chance the ensemble’s majority answer is correct. These voting mechanisms put equal vote on each output. An alternative is to use a voting average or LLM as a judge, where a strong model (or a separate verification model) can be used to pick the best answer among the generations \citep{kim2023prometheus, shridhar2023screws, wang2023math}. ``Think Twice'' \cite{li2024more} is such a framework where the LLM generates multiple answers and also reflects with justifications for each. These justifications are then aggregated to estimate which answer is most likely correct. Essentially, the model is asked to critique or explain each candidate answer, and those explanations help identify the most trustworthy answer. LLM-Debate \cite{du2023improving} is another example where several models ``debate'' an answer and another model (or even the ensemble of arguments) decides the winner, leading to improved performance. Our proposed approach uses a series of LLMs as independent judge, each providing a confidence score for a particular metric, and the final score is an aggregator over all the metrics. This saves multiple generations from the LLMs and the cost is based on the judges instead of the models, which in most of the cases is much cheaper. 

\paragraph{Logits based Fusion}
Another alternative is to combine the probability distributions or logits produced by different LLMs. Instead of only looking at final answers, a fine-grained merging of model outputs at each generation step can be performed. However, the challenge here is that different LLMs have different vocabularies and token representations, making direct averaging of their predicted next-token probabilities impossible if tokens don’t align. DeePEn \cite{huang2024enabling} addresses this by mapping each model’s probability distribution into a universal relative representation space, where token probabilities are compared in terms of rank or relative likelihood. The fused distribution is then mapped back to one model’s token space to pick the next word. LLM-Blender \cite{jiang2023llm} goes one step further and takes a two-step approach: first, a PairRanker model learns to rank multiple candidate answers (through pairwise comparisons), then a Generative Fusion model merges the top answers into one output. However, this is a supervised method, and the ranker needs to be trained, adding additional data and computation overhead. Our approach uses judges over the full generation as for complex tasks like reasoning or long-form generations, judging based on logprobs over a token or even sentence level is not optimal, as the models are often biased towards their own generations \cite{kadavath2022language}.

\paragraph{Task Decomposition based Fusion}
Finally, some other approaches include a cascading approach (Chain of Experts, or CoE) where different models handle different aspects of a single query \cite{xiao2024chainofexperts}. For instance, to solve a math word problem, one could use a specialized parser model to translate the problem into equations, then a math solver model (or tool) to compute the answer, then a language model to phrase the answer in words. Or a query is decomposed into simpler subproblems and each subproblem are solved iteratively \citep{shridhar-etal-2022-automatic, zhou2022least}. Each model’s output feeds into the next – effectively fusing their capabilities in a pipeline rather than merging parallel outputs. However, these approaches are sequential, often leading to very slow input-output cycle. Our task decomposition allows parallel processing of mutliple tasks, leading to no such delay.

\section{\ema}

Let \(\mathcal{F} = \{F_1, F_2, \ldots, F_n\}\) be a collection of \(n\) foundation models, where each model \(F_i\) can be:
\begin{itemize}
    \item \emph{Closed-source}: Accessible via APIs or services, but not publicly modifiable.
    \item \emph{Open-source}: Model weights are available for fine-tuning or other modifications.
\end{itemize}
We denote by \(\mathbf{x}\) the incoming task or input (e.g., a query, an image, or any data point). If the task is decomposed into subproblems, we write \(\{\mathbf{x}_1, \mathbf{x}_2, \ldots, \mathbf{x}_k\}\). Let \(\mathcal{Y}\) be the output space (e.g., text, labels, or embeddings). Below we present our proposed methodology, \ema.

\subsection{Step 1: Problem Decomposition}
\label{sec:step1-prob-decomp}

This initial stage aims to decompose each user request (which may be a monolithic query or a composite instruction) into more tractable subproblems while simultaneously classifying it across a high-coverage taxonomy. Our overarching goal is to enable downstream routing (Steps~2--4) and confidence modulation (CASCADE signals) to adapt according to the task's domain, complexity, modality, and other salient characteristics.

For each subproblem \(\mathbf{x}_j\), we specify task-specific constraints \(\mathcal{C}\), such as \emph{accuracy} thresholds, \emph{latency} requirements, \emph{cost} considerations, or required model \emph{capabilities} (e.g., multi-modal support, structured output, spatial reasoning, etc.). Note that a user can specify their own constraints that they want overall.

\subsection{Step 2: Taxonomy-Based Classification}
\label{sec:taxonomy-routing}

\begin{wrapfigure}{r}{0.5\linewidth}
    \centering
    \vspace{-3.6em} %
    \includegraphics[width=\linewidth]{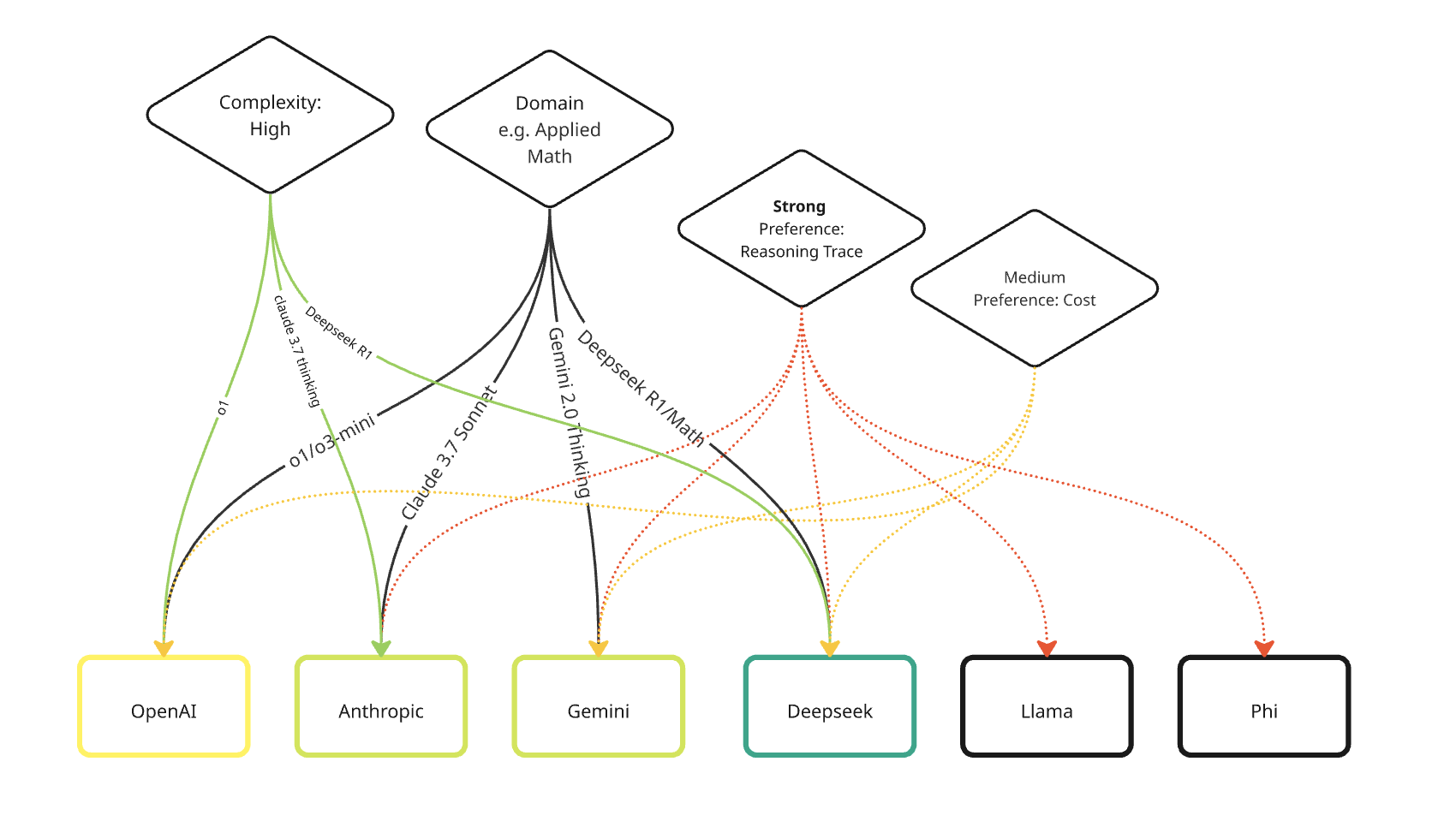}
    \caption{Taxonomy-based routing process.}
    \label{fig:taxonomy-router}
\end{wrapfigure}

In order to route a query \(\mathbf{x}\)---potentially decomposed into subproblems \(\{\mathbf{x}_1, \ldots, \mathbf{x}_k\}\)---we introduce a taxonomy-based classification mechanism that assigns each subproblem \(\mathbf{x}_j\) to one or more high-level categories. These categories reflect key attributes such as task type, reasoning complexity, domain constraints, and input/output formats. 

This classification enables fast routing to a subset of foundation models (see Algorithm~\ref{alg:taxonomy-router}).

\paragraph{Taxonomy Categories.}
For brevity, we present the taxonomy dimensions in a compressed table (Table~\ref{tab:taxonomy-categories}). Each subproblem \(\mathbf{x}_j\) may have multiple labels across dimensions, yielding a multi-label vector \(T(\mathbf{x}_j)\).

\begin{table}[ht]
\centering
\small
\renewcommand{\arraystretch}{1.1}
\setlength{\tabcolsep}{4pt}
\caption{Key taxonomy dimensions with example labels.}
\label{tab:taxonomy-categories}
\resizebox{\linewidth}{!}{%
\begin{tabular}{l|l|l}
\hline
\textbf{Dimension} 
& \textbf{Example Labels}
& \textbf{Description} \\
\hline
\textbf{Task Group} 
& \texttt{instruction\_following}, \texttt{knowledge\_retrieval}, \texttt{analytical\_reasoning} 
& High-level function or objective. \\
\textbf{Reasoning Type} 
& \texttt{single\_step}, \texttt{multi\_hop}, \texttt{chain\_of\_thought}
& Depth and style of inference.\\
\textbf{I/O Format} 
& \texttt{plain\_text}, \texttt{json}, \texttt{program\_code} 
& Input format or required output structure.\\
\textbf{Domain} 
& \texttt{medical}, \texttt{legal}, \texttt{finance}
& Specialized subject area.\\
\textbf{Complexity} 
& \texttt{low}, \texttt{medium}, \texttt{high}
& Overall difficulty (reasoning steps, knowledge required).\\
\hline
\end{tabular}%
}
\end{table}

\paragraph{Slow vs. Fast Inference.}
We present two complementary taxonomy-based classification approaches in our work:

\begin{enumerate}[label=(\roman*)]
    \item \textbf{Slow (LLM-based) Classifier}:
    A high-capacity model (e.g., GPT-4 \cite{achiam2023gpt}) parses \(\mathbf{x}_j\) and outputs a probability distribution over possible taxonomy labels. Denote these probabilities by
    \[
      p_{\mathrm{slow}}\bigl(c \mid \mathbf{x}_j\bigr),
    \]
    for each label \(c\). This method is accurate but expensive and slower.
    
    \item \textbf{Fast (Embedding-based) Classifier}:
    For real-time or cost-sensitive scenarios, we embed \(\mathbf{x}_j\) into a vector space. Let \(\mathbf{v}_j = \text{Embed}(\mathbf{x}_j)\in\mathbb{R}^d\). We also store reference embeddings \(\{\mathbf{u}_c\in\mathbb{R}^d\}\) for each label \(c\), or for a set of reference examples for label \(c\). We define a distance-based assignment:
    \[
      d\bigl(\mathbf{v}_j, \mathbf{u}_c\bigr) 
      \;=\; \|\mathbf{v}_j - \mathbf{u}_c\|_2,
    \]
    and convert distances to the label probabilities
    \[
      p_{\mathrm{fast}}\bigl(c \mid \mathbf{x}_j\bigr) 
      \;=\; \frac{ \exp\bigl(-\alpha \,d(\mathbf{v}_j,\mathbf{u}_c)\bigr) }
                  { \sum_{c'} \exp\bigl(-\alpha \,d(\mathbf{v}_j,\mathbf{u}_{c'})\bigr)},
    \]
    where \(\alpha>0\) is a temperature-like scale factor.  %
\end{enumerate}

\paragraph{Fusion of Slow \& Fast Classifications.}
We fuse the two distributions as:
\begin{equation}
\label{eq:taxonomy-fusion}
p_{\mathrm{fusion}}(c \mid \mathbf{x}_j)
\;=\;
\lambda \,p_{\mathrm{slow}}(c \mid \mathbf{x}_j)
\;+\;
(1 - \lambda)\, p_{\mathrm{fast}}(c \mid \mathbf{x}_j),
\end{equation}
where \(\lambda \in [0,1]\) is a hyperparameter either set by the user or decided based on the cost. We then select the final taxonomy labels for \(\mathbf{x}_j\) by thresholding:
\[
T(\mathbf{x}_j) \;=\; 
\bigl\{
  c \;\mid\; p_{\mathrm{fusion}}(c \mid \mathbf{x}_j) \;\ge\; \tau_{\text{label}}
\bigr\},
\]
or by taking top-$k$ labels in descending probability.

\paragraph{Routing Using Taxonomy Scores.}
Once the final labels \(T(\mathbf{x}_j)\) are assigned, we compute a “suitability” function \(\Phi(F_i, T(\mathbf{x}_j))\) for each model \(F_i \in \mathcal{F}\). In our case, \(\Phi\) is a simple binary check (for example, does \(F_i\) claim strong performance on a given category). We then route:
\[
\text{Route}_{\text{taxonomy}}(\mathbf{x}_j)
=
\bigl\{
  F_i \in \mathcal{F} 
  \;\mid\;
  \Phi\bigl(F_i, T(\mathbf{x}_j)\bigr) \;\ge\; \tau_c
\bigr\}.
\]
If this subset is non-empty and we have high confidence in the taxonomy assignment (e.g., no ambiguous domain or reasoning requirement), we skip the Learned Router (Step 3) and directly select from \(\text{Route}_{\text{taxonomy}}(\mathbf{x}_j)\). Otherwise, we defer to the next step.

\begin{algorithm}[ht]
\caption{Taxonomy-Based Routing with Slow \& Fast Classifiers}
\label{alg:taxonomy-router}
\begin{algorithmic}[1]
\Require Subproblem $\mathbf{x}_j$, 
       LLM-based classifier ($p_{\mathrm{slow}}$), 
       Embedding-based classifier ($p_{\mathrm{fast}}$), 
       Thresholds $\tau_{\text{label}}, \tau_c, \lambda$, 
       Suitability function $\Phi$, 
       Model collection $\mathcal{F}$
\Ensure Candidate set $S_{\text{tax}} \subseteq \mathcal{F}$
\Statex
\State \textbf{Step 1: Slow Classification} 
       \Comment{LLM-based (if budget/time allows)}
\State $\{p_{\mathrm{slow}}(c \mid \mathbf{x}_j)\}_{c\in\mathcal{C}} \gets \mathrm{LLMClassifier}(\mathbf{x}_j)$
\State \textbf{Step 2: Fast Classification} 
       \Comment{Embedding-based / vector space lookup}
\State $\mathbf{v}_j \gets \mathrm{Embed}(\mathbf{x}_j)$
\For{each label $c$ in candidate label set $\mathcal{C}$}
    \State $\mathbf{u}_c \gets \mathrm{RefEmbed}(c)$ \Comment{Pre-stored label embedding}
    \State $d_c \gets \|\mathbf{v}_j - \mathbf{u}_c\|_2$ 
       \Comment{Distance in embedding space}
\EndFor
\State Convert $\{d_c\}$ to $p_{\mathrm{fast}}(c \mid \mathbf{x}_j)$ by using a softmax function
\State \textbf{Step 3: Probability Fusion}
\For{each label $c$ in $\mathcal{C}$}
    \State $p_{\mathrm{fusion}}(c \mid \mathbf{x}_j)
           \;\gets\;
           \lambda \,p_{\mathrm{slow}}(c \mid \mathbf{x}_j)
           + (1-\lambda)\,p_{\mathrm{fast}}(c \mid \mathbf{x}_j)$
\EndFor
\State \textbf{Step 4: Final Label Selection}
\State $T(\mathbf{x}_j) \gets \bigl\{ c : p_{\mathrm{fusion}}(c \mid \mathbf{x}_j) \ge \tau_{\text{label}} \bigr\}$
\State \textbf{Step 5: Model Suitability}
\State $\text{Route}_{\text{taxonomy}}(\mathbf{x}_j) \gets \bigl\{ F_i \in \mathcal{F} \mid \Phi(F_i, T(\mathbf{x}_j)) \ge \tau_c \bigr\}$
\State \Return $\text{Route}_{\text{taxonomy}}(\mathbf{x}_j)$
\end{algorithmic}
\end{algorithm}

\subsection{Step 3: Learned Router}
\label{sec:learned-router}

While taxonomy-based routing covers queries in “high-confidence” or well-defined categories, many tasks remain ambiguous or out-of-distribution. For these cases, we employ a \emph{learned router} to predict which model(s) in \(\mathcal{F}\) will perform best under the user’s constraints.

\paragraph{Learned Router Formulation.}
We treat model selection as a multi-output regression problem. For each subproblem \(\mathbf{x}_j\), our goal is to predict an expected performance score \(s(F_i, \mathbf{x}_j)\) for each foundation model \(F_i \in \mathcal{F}\). The learned router then selects up to \(k\) models to maximize the overall performance:
\[
  \text{Route}_{\text{learned}}(\mathbf{x}_j)
  \;=\;
  \underset{S \subseteq \{1,\dots,n\},\, |S|\le k}{\arg\max} 
  \sum_{i \in S} s\bigl(F_i, \mathbf{x}_j\bigr)
  \quad
  \text{subject to}\;\; \mathcal{C},
\]
where \(\mathcal{C}\) encapsulates feasibility (e.g., cost or domain restrictions). 

\paragraph{Router Model Architecture.}
A parametric function \(R_\theta\) outputs \(\hat{s}_{F_i,\mathbf{x}_j}\) (an estimate of \(s(F_i,\mathbf{x}_j)\)) for each model \(F_i\). Concretely, let
\[
h_{\mathbf{x}} \;=\; \mathrm{Enc}(\mathbf{x}), 
\quad
f_{\mathrm{tax}} \;=\; \mathrm{Embed}\bigl(T(\mathbf{x})\bigr),
\quad
h_{\mathrm{combined}} \;=\; \mathrm{LayerNorm}\bigl(\,[\,h_{\mathbf{x}};\,f_{\mathrm{tax}}\,]\bigr).
\]
Then:
\[
\hat{s}_{F_i,\mathbf{x}}
\;=\;
w_i^{\top}\,h_{\mathrm{combined}} + b_i,
\]
which is normalized (e.g., via \(z\)-scoring) to accommodate per-model scale differences. The model is trained to minimize the mean squared error (MSE) between \(\hat{s}_{\mathrm{norm}}(F_i,\mathbf{x}_j;\theta)\) and the ground-truth normalized performance \(s_{\mathrm{norm}}(F_i,\mathbf{x}_j)\):
\[
\min_{\theta}\; \frac{1}{N \,\cdot\, |\mathcal{F}|}
\sum_{j=1}^{N}
\sum_{i=1}^{|\mathcal{F}|}
\Bigl(
\hat{s}_{\mathrm{norm}}(F_i,\mathbf{x}_j;\theta)
-
s_{\mathrm{norm}}(F_i,\mathbf{x}_j)
\Bigr)^2.
\]
At inference, we denormalize the prediction using each model’s \(\mu_{F_i}, \sigma_{F_i}\) from training:
\[
\hat{s}_{\mathrm{raw}}(F_i,\mathbf{x})
=
\hat{s}_{\mathrm{norm}}(F_i,\mathbf{x}) \cdot \sigma_{F_i} 
\;+\;
\mu_{F_i},
\]
and pick the top-\(k\) models that pass user constraints \(\mathcal{C}\).

\begin{algorithm}[ht]
\caption{Hybrid Routing Algorithm for LLM Selection (Combining Taxonomy + Learned)}
\label{alg:hybrid_routing}
\begin{algorithmic}[1]
\Require Input subproblem $\mathbf{x}$, Foundation models $\mathcal{F}$, Taxonomy classifier $T(\mathbf{x})$ with suitability function $\Phi(\cdot)$, Learned router $R(\mathbf{x})$ that predicts model performance scores, Constraints $\mathcal{C}$ (e.g., cost, domain restrictions), Subset sizes $k_{\text{tax}}$, $k_{\text{lr}}$ for taxonomy and learned outputs, Maximum final selection size $n$ (where $n \leq k_{\text{tax}} + k_{\text{lr}}$).

\Ensure Selected model(s) $S_{\mathrm{hybrid}} \subseteq \mathcal{F}$ of size $\le n$.
\Statex

\State \textbf{(1) Taxonomy Router:} 
\Statex \hspace*{1em} Run taxonomy-based classification to get categories $C_{\mathbf{x}} \gets T(\mathbf{x})$.
\For{each model $F_i \in \mathcal{F}$}
    \State $\phi_i \gets \Phi\bigl(F_i, C_{\mathbf{x}}\bigr)$ 
       \Comment{Taxonomy suitability score}
\EndFor
\State $S_{\mathrm{tax}} \gets \text{TopK}\bigl(\{\phi_i\}, k_{\text{tax}}\bigr) \cap \text{FilterByConstraints}(\mathcal{F}, \mathcal{C})$

\vspace{0.5em}
\State \textbf{(2) Learned Router:} 
\Statex \hspace*{1em} Predict performance scores for each model.
\For{each model $F_i \in \mathcal{F}$}
    \State $s_i \gets R\bigl(\mathbf{x}, C_{\mathbf{x}}\bigr)[i]$
       \Comment{Predicted performance score for $F_i$}
\EndFor
\State $S_{\mathrm{learned}} \gets \text{TopK}\bigl(\{s_i\}, k_{\text{lr}}\bigr) \cap \text{FilterByConstraints}(\mathcal{F}, \mathcal{C})$

\vspace{0.5em}
\State \textbf{(3) Combine \& Final Selection:}
\State $S_{\mathrm{union}} \gets S_{\mathrm{tax}} \;\cup\; S_{\mathrm{learned}}$
\If{$|S_{\mathrm{union}}| \le n$}
    \State $S_{\mathrm{hybrid}} \gets S_{\mathrm{union}}$
\Else
    \Comment{Pick best $n$ models (e.g. by $s_i$) from the union}
    \State $S_{\mathrm{hybrid}} \gets \text{TopN}\bigl(\{s_i : i \in S_{\mathrm{union}}\}, n\bigr)$
\EndIf
\State \Return $S_{\mathrm{hybrid}}$
\end{algorithmic}
\end{algorithm}

\subsubsection{Hybrid Routing (Taxonomy + Learned)}
\label{sec:hybrid-routing}

Although taxonomy-based classification (Step~2) efficiently handles tasks that fall into well-understood categories, a pure taxonomy-based approach may miss opportunities or fail in cases where the taxonomy is incomplete or the classification confidence is only moderate. Conversely, a purely learned router (Step~3) can be more flexible but may incur higher computational costs for every query.

We propose a \emph{hybrid} selection strategy that leverages both:
\begin{itemize}[leftmargin=2em]
    \item A \textbf{Taxonomy Router} that returns a set of models \(S_{\mathrm{tax}}\) believed suitable based on category alignment.
    \item A \textbf{Learned Router} that selects a set of models \(S_{\mathrm{learned}}\) by predicting performance scores.
\end{itemize}
We then \emph{combine} these two subsets to produce a final collection of up to \(n\) models for downstream execution. This design ensures that well-understood (taxonomically clear) cases remain covered by specialized models, while ambiguous or novel tasks can still draw from the learned router’s data-driven predictions.

Algorithm~\ref{alg:hybrid_routing} outlines the hybrid approach. First, we invoke the \textbf{taxonomy router} to obtain a subset of candidate models \(S_{\mathrm{tax}}\). Next, we run the \textbf{learned router} to generate performance scores and extract another subset \(S_{\mathrm{learned}}\). Finally, we \emph{merge} the two sets and, if necessary, pick the top-$n$ models according to the learned scores (or other priority criteria) to form our final selection \(S_{\mathrm{hybrid}}\).

\subsection{Step 4: Cascading Routing}
Let \(\mathcal{S}^*_j\) be the final subset of candidate models obtained from the taxonomy or learned router for subproblem \(\mathbf{x}_j\). In many scenarios, we adopt a \emph{cascading} strategy to balance cost and performance:
\begin{enumerate}
    \item Sort the models in \(\mathcal{S}^*_j\) in increasing order of the user's choice of criteria (e.g., from cheapest to most expensive).
    \item For each model \(F_{k}\) in that sorted list:
    \begin{enumerate}
        \item Generate an output \(y_{k} = F_{k}(\mathbf{x}_j)\).
        \item \emph{Employ a judge} to evaluate \(y_{k}\) and produce a confidence score \(\pi_{k}\). In our framework, this judge is realized by the \emph{CASCADE} algorithm (detailed below), which combines multiple signals---including an LLM-based evaluation---into a single confidence metric.
        \item If \(\pi_{k} \ge \delta\) (a threshold), stop and accept \(y_{k}\).
        \item Otherwise, escalate to the next (more expensive or more capable) model in \(\mathcal{S}^*_j\).
    \end{enumerate}
    \item If none of the models in \(\mathcal{S}^*_j\) produce a confidence above \(\delta\) after trying all, return a fallback (e.g., “no confident solution” or a fixed large model).
\end{enumerate}
We denote the final model chosen for subproblem \(\mathbf{x}_j\) (or the final output produced) by 
\[
F_{\text{cascade}}(\mathbf{x}_j).
\]

\subsubsection{CASCADE: A Context-Aware Signal Combination Algorithm}
At the heart of our cascading approach is the CASCADE (Context-Aware Signal Combination And Deferral Evaluation) algorithm, which serves as our \emph{judging criteria}. CASCADE makes deferral decisions based on multiple complementary confidence signals. While recent approaches have explored learning-based methods for weighting these signals \cite{dekoninck2024unified}, we adopt a principled, deterministic algorithm that eliminates the need for continuous weight optimization while achieving comparable performance.

CASCADE leverages five distinct confidence signals:
\begin{enumerate}
    \item $S_L(y_i, x)$: \textbf{Logit-based confidence} derived from model probabilities, which provides fast computation but may be vulnerable to model overconfidence.
    \item $S_S(y_i, x)$: \textbf{Self-reported confidence} solicited directly from the model during generation (usually after each substantial sentence or chunk), effective at identifying knowledge boundaries.
    \item $S_R(y_i, x)$: \textbf{Reward model score} evaluating response quality, which captures stylistic elements, coherence, and overall quality.
    \item $S_D(y_i, x)$: \textbf{Domain-specific verification score} that provides high-precision evaluation for specialized domains.
    \item $S_J(y_i, x)$: \textbf{LLM judge evaluation} offering comprehensive assessment through specialized models.
\end{enumerate}

The key insight of CASCADE is that these signals have varying reliability across different query domains. Rather than using a one-size-fits-all approach, CASCADE uses domain-specific static weighting vectors. 

We employ judging the model output from the cascading router by setting all confidence scores ($S_i$) to 0. We start by classifying the domain of the query based on a trained binary classifier and apply domain-specific verification. We also check if the logits-based confidence is available for the selected model or not. If not, we use a self-reported confidence of the output following \cite{wei2024measuring}. Finally, we apply a reward model (described below in details) to the output and calculate a weighted confidence $S_{\text{combined}}$. We check if the weighted confidence measure if borderline or not by defining borderline parameters range $(\tau_{\text{borderline\_low}} \le S_{\text{combined}} \le \tau_{\text{borderline\_high}}$). We invoke an LLM as a judge if borderline scores are present as an additional confidence estimation parameter. A detailed analysis of the CASCADE algorithm is presented in \autoref{alg:CASCADE}.

\begin{algorithm}[t]
\caption{CASCADE Algorithm}
\label{alg:CASCADE}
\begin{algorithmic}[1]
\State \textbf{Step 1:} Classify the domain of the query.
\State \textbf{Step 2:} Evaluate the logit-based confidence $S_L$ and apply Fast Classification from Taxonomy Routing :
    \begin{enumerate}[label=(\alph*)]
        \item If $S_L > \tau_{\text{high}}$  and the domain is not sensitive, \textbf{accept} the response immediately.
        \item If $S_L < \tau_{\text{low}}$, \textbf{defer} immediately.
    \end{enumerate}
\State \textbf{Step 3:} For specialized domains, apply domain-specific verification:
    \begin{enumerate}[label=(\alph*)]
        \item If $S_D < \tau_{\text{domain}}$, \textbf{defer} due to verification failure.
    \end{enumerate}
\State \textbf{Step 4:} Extract self-reported confidence $S_S$ (if present):
    \begin{enumerate}[label=(\alph*)]
        \item If a knowledge boundary is detected ($\min(S_S) < \tau_{\text{knowledge}}$), \textbf{defer} unless this is the final model.
    \end{enumerate}
\State \textbf{Step 5:} Evaluate the reward model score $S_R$.
\State \textbf{Step 6:} Compute the weighted combination:
    \[
      S_{\text{combined}} 
      \;=\;
      \mathbf{w}[1] \cdot S_L
      \;+\;
      \mathbf{w}[2] \cdot S_S
      \;+\;
      \mathbf{w}[3] \cdot S_R
      \;+\;
      \mathbf{w}[4] \cdot S_D.
    \]
\State \textbf{Step 8:} For borderline cases 
    ($\tau_{\text{borderline\_low}} \le S_{\text{combined}} \le \tau_{\text{borderline\_high}}$),
    \textbf{invoke the LLM judge}:
    \begin{enumerate}[label=(\alph*)]
        \item If $S_J < 0.5$, \textbf{defer}; otherwise, \textbf{accept}.
    \end{enumerate}
\State \textbf{Step 9:} If not borderline:
    \begin{enumerate}[label=(\alph*)]
        \item If $S_{\text{combined}} < 0.5$, \textbf{defer}.
        \item If $S_{\text{combined}} \ge 0.5$, \textbf{accept}.
    \end{enumerate}
\end{algorithmic}
\end{algorithm}

Finally, the CASCADE output for a model’s prediction \(y_k\) is:
\[
    \pi_k \;=\; S_{\text{CASCADE}}(y_k, x),
\]
where \(S_{\text{CASCADE}}\) is computed by the above procedure, including the LLM judge only as needed.

\paragraph{Reward Model}
The reward model provides a crucial quality signal within CASCADE, capturing stylistic, coherence, and content aspects that may not be reflected in other confidence metrics.

Our reward model $\mathcal{R}_\theta$ employs a dual-encoder architecture that separately processes the query and response before computing a quality score:
\[
\mathcal{R}_\theta(y, x) = \sigma\bigl(f_\theta(h_x, h_y)\bigr),
\]
where $h_x$ and $h_y$ are encoded representations of the query and response, $f_\theta$ is a scoring function, and $\sigma$ is a sigmoid activation that maps scores to $[0,1]$.

\paragraph{LLM as a Judge}
The LLM judge component represents the most sophisticated evaluation signal in CASCADE. Unlike traditional approaches that employ fixed evaluation criteria, our system generates query-specific evaluation rubrics and applies them through specialized judge models.

Our judge system employs a two-stage architecture:
\[
\mathcal{J}(y, x) = \mathcal{J}_{\text{eval}}\bigl(y, x, r\bigr),
\]
where $r = \mathcal{J}_{\text{rubric}}(x)$ is the dynamically generated evaluation rubric.

The rubric generator produces a structured rubric containing evaluation dimensions, dimension weights, and scoring criteria tailored to the specific query type. The evaluation model then analyzes the response according to this rubric, producing dimension-specific scores, an overall quality score, and a confidence estimate.

\subsection{Step5: Task Decomposition Fusion }
If a task is decomposed into multiple subproblems in Step 1, each subproblem can be routed through the cascade independently, possibly with parallel or sequential executions. Specifically, when the final subset \(\mathcal{S}^*_j\) for subproblem \(\mathbf{x}_j\) is used in \textbf{parallel} rather than a strict cascade---or if multiple models are run to capture different aspects of the problem---\ema\ fuses their outputs. Concretely, given the input \(\mathbf{x}_j\) and each selected model \(F_i \in \mathcal{S}^*_j\), we obtain individual outputs
\[
    y_i \;=\; F_i(\mathbf{x}_j),
    \quad\text{for each}\quad i \in \mathcal{S}^*_j.
\]
We then define a fusion function
\[
    f_{\text{fusion}} : \bigl(\mathcal{Y}\bigr)^{|\mathcal{S}^*_j|} \;\to\; \mathcal{Y},
\]
which aggregates or integrates the multiple outputs \(\{y_i \mid i \in \mathcal{S}^*_j\}\) into a single final output
\[
    \mathbf{y}_j \;=\; f_{\text{fusion}}\bigl(y_i \;\big|\; i \in \mathcal{S}^*_j\bigr).
\]
Any fused result can again be assessed by the CASCADE pipeline for quality assurance.

\section{Experiments}

\subsection{Dataset}
\label{sec:dataset}

\begin{table}[ht!]
\centering
\resizebox{\textwidth}{!}{%
\begin{tabular}{lrrrr}
\toprule
\multirow{2}{*}{\textbf{Category}} & \multicolumn{2}{c}{\textbf{Training Set (N = 15,023)}} & \multicolumn{2}{c}{\textbf{Evaluation Set (N = 1,567)}} \\
\cmidrule(lr){2-3} \cmidrule(lr){4-5}
& \textbf{\# Samples} & \textbf{Dist. (\%)} & \textbf{\# Samples} & \textbf{Dist. (\%)} \\
\midrule
\multicolumn{5}{c}{\textbf{Instruction Following (Easy)}} \\
\midrule
\quad IFEvaL \cite{ifeval}                          & 793  & 5.27\%  & 100  & 6.38\% \\
\quad [Enterprise] Executive Summary Generation      & 231  & 1.54\%  & 50   & 3.19\% \\
\quad [Enterprise] Multi-constraint Content Reformatting & 200  & 1.33\%  & 50   & 3.19\% \\
\quad [Enterprise] Conversational State Tracking & 200  & 1.33\%  & 50   & 3.19\% \\
\quad [Enterprise] Standardized Output Structuring   & 200  & 1.33\%  & 50   & 3.19\% \\
\quad [Enterprise] Support Request Classification    & 200  & 1.33\%  & 50   & 3.19\% \\
\midrule
\multicolumn{5}{c}{\textbf{Instruction Following (Hard)}} \\
\midrule
\quad Tulu3 \cite{tulu3-sft}                        & 1004 & 6.68\%  & 100  & 6.38\% \\
\quad [Enterprise] RFP Response Enhancement          & 493  & 3.28\%  & 50   & 3.19\% \\
\quad [Enterprise] CX Ticket/Conversation Understanding & 200 & 1.33\% & 50   & 3.19\% \\
\quad [Enterprise] Proposal Document Refinement      & 200  & 1.33\%  & 50   & 3.19\% \\
\midrule
\multicolumn{5}{c}{\textbf{Knowledge Context Reasoning (Easy)}} \\
\midrule
\quad MMLU \cite{mmlu}                              & 982  & 6.53\%  & 70   & 4.47\% \\
\quad MMLU Pro                               & 687  & 4.57\%  & 49   & 3.13\% \\
\quad [Enterprise] Multi-modal Understanding \& Reranking & 200 & 1.33\% & 51 & 3.25\% \\
\quad [Enterprise] Search Relevance \& Reranking     & 200  & 1.33\%  & 48   & 3.06\% \\
\quad [Enterprise] Context-aware Query Suggestion    & 200  & 1.33\%  & 50   & 3.19\% \\
\quad [Enterprise] Hybrid Context Q\&A               & 200  & 1.33\%  & 50   & 3.19\% \\
\midrule
\multicolumn{5}{c}{\textbf{Knowledge Context Reasoning (Hard)}} \\
\midrule
\quad [Enterprise] RFP Generation                    & 200  & 1.33\%  & 50   & 3.19\% \\
\quad [Enterprise] Complex CX Resolution             & 200  & 1.33\%  & 50   & 3.19\% \\
\quad [Enterprise] Long-Context Understanding \& Generation & 200 & 1.33\% & 49 & 3.13\% \\
\quad [Enterprise] Hierarchical Proposal Assistance  & 200  & 1.33\%  & 50   & 3.19\% \\
\midrule
\multicolumn{5}{c}{\textbf{Quantitative Analytical Reasoning (Easy)}} \\
\midrule
\quad GSM8K \cite{cobbe2021gsm8k}                   & 1272 & 8.46\%  & 100  & 6.38\% \\
\quad ARC \cite{allenai_arc}                        & 539  & 3.58\%  & 100  & 6.38\% \\
\midrule
\multicolumn{5}{c}{\textbf{Quantitative Analytical Reasoning (Hard)}} \\
\midrule
\quad HLE \cite{phan2025humanity}                   & ---  & ---     & 100  & 6.38\% \\
\midrule
\multicolumn{5}{c}{\textbf{Code Generation}} \\
\midrule
\quad MBPP \cite{mbpp}                              & ---  & ---     & 100  & 6.38\% \\
\quad [Enterprise] Technical Query Formulation       & 240  & 1.60\%  & 50   & 3.19\% \\
\midrule
\quad Other Enterprise Tasks                         & 5682 & 37.82\% & 0    & 0.00\% \\
\bottomrule
\end{tabular}
}
\caption{Comparison of the training and evaluation datasets organized by task categories. The training dataset (15,023 samples) contains a broader range of tasks including many enterprise-specific tasks summarized in "Other Enterprise Tasks", while the evaluation set (1,567 samples) maintains a more balanced distribution across open-source benchmarks and enterprise tasks. All enterprise tasks are labeled with "[Enterprise]" prefix to clearly differentiate them from open-source benchmarks.}
\label{tab:combined-dataset}
\end{table}

We constructed training and testing datasets to evaluate the effectiveness of our fusion approach across various tasks, including mathematical reasoning, question answering, and constrained instruction-following tasks, in both multiple-choice and free-response formats. Our training dataset combines open-source datasets with proprietary data, whereas evaluation is performed on open-source benchmarks. \autoref{tab:combined-dataset} details the distribution of both training and testing data. A comprehensive distribution is presented in \autoref{sec:full-dataset-table}.

\paragraph{Data Sampling Methodology.}
The training set is intentionally diverse, covering a broad spectrum of tasks, while the testing set consists of diverse and particularly challenging problems, such as those from the Humanity Last Exam \cite{phan2025humanity}. We selectively sample the most difficult tasks (e.g., long reasoning chains, elaborate instructions) to build a training corpus that stresses corner cases and diverse reasoning patterns. For instance, Tulu3 SFT Personas is filtered to instances with constraint length 3, and IFEval-like sets require responses with at least 10 sentences. When no inherent constraints are available, we sample randomly from the hardest quartile (based on problem metadata or pilot study performance) to keep the training data tractable in size but rich in complexity.

Common benchmarks like GSM8K \cite{cobbe2021gsm8k}, ARC \cite{allenai_arc}, and SFT Personas \cite{tulu3-sft} are represented in both datasets. To reduce the training dataset size, we randomly sampled the most challenging problems, applying constraints where applicable (e.g., setting the constraint length to 3 for Tulu3 SFT and requiring a minimum of 10 sentences for IFEval-like datasets). When no constraints were available, random sampling was used.

\begin{table}[ht]
\centering
\small
\renewcommand{\arraystretch}{1.2}
\caption{Dimension-level descriptors used for multi-dimensional scoring.}
\label{tab:judge-dimensions}
\begin{tabular}{l|p{10.5cm}}
\hline
\textbf{Dimension} & \textbf{Description} \\
\hline
\textbf{Instruction Following} 
  & Accuracy in adhering to user directives or system constraints.\\
\textbf{Factual Correctness} 
  & Congruence with verified facts, especially critical for knowledge-intensive tasks.\\
\textbf{Reasoning Quality} 
  & Logical coherence and depth of the chain of thought.\\
\textbf{Completeness} 
  & Extent to which the response addresses all relevant facets of the query.\\
\textbf{Clarity \& Organization} 
  & Readability and structure of the provided solution.\\
\textbf{Relevance} 
  & Alignment with the user's intended topic or requested content.\\
\textbf{Helpfulness} 
  & Practical utility of the response for the user's goal.\\
\hline
\end{tabular}
\end{table}

\paragraph{Judge data}

Once we have the training data split ready, we create the dataset to train the routers and the judges used in our work. A core objective of our dataset curation is to create ``oracle-like'' labels that accurately reflect response quality across several dimensions. Empirically, single-metric evaluations (e.g., simple correctness flags) can miss critical aspects such as clarity, completeness, or instruction-following fidelity. We therefore adopt a multi-dimensional framework as defined in Table~\ref{tab:judge-dimensions} (and a detailed description provided in Appendix \autoref{app:eval-framework}) and using multiple LLMs judges, we curate our judge-focused dataset through the following steps:

\begin{enumerate}
    \item \textbf{Define Multi-Dimensional Framework.}
    We establish the dimensions for quality assessment (\emph{Instruction Following}, \emph{Factual Correctness}, \emph{Reasoning Quality}, etc.) as shown in Table~\ref{tab:judge-dimensions}. This ensures we capture nuanced aspects of response quality beyond simple ``correctness'' flags.
    
    \item \textbf{Collect Candidate Responses.}
    For each query (instruction or question) in our dataset, we gather \emph{candidate responses} produced by all LLMs used in this work. These responses exhibit a broad spectrum of quality, from trivial to highly complex failures, enabling the judges to learn robustly.

    \item \textbf{Run Multiple Judge Models.}
    We employ two distinct LLM-based judges---(\emph{o3-mini} \cite{achiam2023gpt} and \emph{Claude~3.7 Sonnet (Reasoning)}) \cite{anthropic2024claude3}---to score each query-response pair across all predefined dimensions. This \emph{Multiple Judge Models} step reduces correlated biases since the judges differ in architecture and pretraining.

    \item \textbf{Dimension-Averaged Scoring.}
    Each judge is run multiple times on the same pair with minor prompt variations. We \emph{average} their dimension-level scores to mitigate variance and produce more stable annotations. For example, if judge~A reports \(\texttt{factual\_correctness}=3.0\) and judge~B reports \(2.7\), we record \(2.85\) for that dimension.

    \item \textbf{Human Expert Review for Disagreements.}
    Whenever the two judges diverge by more than $2$ points on the overall (1--5) scale, we \emph{flag} the sample for expert human review. This step ensures that specialized or domain-specific queries that cause large disagreements are resolved accurately. In practice, fewer than 5\% of samples require such manual adjudication.
\end{enumerate}

A comprehensive discussion is provided in Appendix \autoref{app:multi-judge-system}. 
\subsubsection*{Final Data Representation}

Each curated data instance thus consists of:
\begin{enumerate}
  \item The \emph{query} (an instruction or question).
  \item The \emph{candidate response} from a smaller or intermediate LLM model in our cascade.
  \item \emph{Dimension-level annotations} (e.g., \texttt{factual\_correctness}$=2.8$, \texttt{reasoning\_quality}$=2.4$, etc.), averaged across multiple runs and across both judges.
  \item \emph{Overall quality scores} on a 1--5 scale, again aggregated unless flagged for review.
\end{enumerate}

\subsection{Models}

\paragraph{Training Details}

For the embedding-based classifier used in taxonomy routing, we embed the inputs into a vector using sentence transformer \cite{thakur-2020-AugSBERT}. We used the ModernBERT \cite{modernbert} as the model. Similarly, ModernBERT was used for the input encoding for the learned router in \autoref{sec:learned-router}. Moreover, AdamW optimizer \citep{loshchilov2017decoupled} with decoupled weight decay was used with a learning rate of $2 \times 10^{-5}$ with linear warmup over the first 10\% of steps. A batch size of 4 was used with early stopping, with patience of 2 evaluation periods (evaluated every 300 steps). Dropout with a 0.1 value was used in both the encoder and regression heads. Weight decay ($1 \times 10^{-2}$) for regularization. The training typically converges within 3-5 epochs, with the validation loss stabilizing after approximately 1,500 optimization steps. Finally, we use 
Skywork-Reward-Llama-3.1-8B-v0.2  \cite{liu2024skywork} as the reward model-based judge in our work. We finetuned it further on a subset of our internal dataset for improved performance. 

\paragraph{Hyper-parameter Setting} For CASCADE algorithm, we set the logit-based confidence thresholds $\tau_{\text{high}}$ as 0.95 and $\tau_{\text{low}}$ as 0.3. For specialized domains, the domain-specific verification threshold $\tau_{\text{domain}}$ is set as 0.7. Borderline detection cases are checked at 0.5. 

\paragraph{Models employed for Routing}
We employed a combination of open-source and closed-source models reputed for state-of-the-art performance across multiple domains. The closed-source models include ChatGPT variants \cite{achiam2023gpt} (\texttt{gpt-3.5-turbo}, \texttt{gpt-4}, \texttt{gpt-4o-mini}, and \texttt{gpt-4o}) alongside their reasoning counterparts (\texttt{O1-mini}, \texttt{O1} and \texttt{O3-mini}), Claude variants \cite{anthropic2024claude3} (\texttt{claude3-haiku}, \texttt{claude3.5-haiku}, \texttt{claude3-opus}, and \texttt{claude3.5-sonnet}), and Gemini variants \cite{team2024gemini} (\texttt{gemini-1.5-pro}, \texttt{gemini-1.5-flash}, \texttt{gemini-2.0-flash-lite}, \texttt{gemini-2.0-flash}, and \texttt{gemini-2.0-flash-thinking}). For open-source alternatives, we utilized Llama models \cite{grattafiori2024llama} (\texttt{llama3.2-90b}, \texttt{llama3.1-405b} and \texttt{llama-3.3-70b}), DeepSeek V3 \cite{liu2024deepseek}, DeepSeek R1 \cite{guo2025deepseekR1} and DeepSeek R1 Distilled Qwen 32B.

\section{Results}

\definecolor{highperf}{RGB}{225,255,225}   %
\definecolor{goodperf}{RGB}{245,255,240}   %
\definecolor{midperf}{RGB}{255,248,225}    %
\definecolor{lowperf}{RGB}{255,235,235}    %

\begin{table}[ht!]
    \centering
    \caption{Summarized Performance Stratification Across Key Taxonomic Categories (see Appendix~\ref{sec:taxonomy-based-routing-performance-expanded} for complete results). Performance values are color-coded from highest (light green) to lowest (light peach).}
    \label{tab:taxonomy-performance}
    \begin{tabular}{p{8cm}cc}
    \toprule
    \textbf{Category} & \textbf{Top-1 Acc.} & \textbf{Top-3 Acc.} \\
    \midrule
    \rowcolor[gray]{0.95} \multicolumn{3}{l}{\textit{Task Groups}} \\
    \quad quantitative\_analytical\_reasoning & \cellcolor{highperf}93.89\% & \cellcolor{highperf}94.66\% \\
    \quad knowledge\_context\_reasoning: RAG & \cellcolor{midperf}88.89\% & \cellcolor{midperf}88.89\% \\
    \quad instruction\_following: multi\_step\_procedure & \cellcolor{midperf}88.69\% & \cellcolor{midperf}90.72\% \\
    \midrule
    \rowcolor[gray]{0.95} \multicolumn{3}{l}{\textit{Reasoning Types}} \\
    \quad arithmetic & \cellcolor{highperf}95.93\% & \cellcolor{highperf}99.19\% \\
    \quad instruction\_analysis & \cellcolor{highperf}93.75\% & \cellcolor{highperf}93.06\% \\
    \quad multi\_step\_reasoning & \cellcolor{midperf}88.38\% & \cellcolor{midperf}90.66\% \\
    \quad causal\_reasoning & \cellcolor{lowperf}82.00\% & \cellcolor{lowperf}86.00\% \\
    \midrule
    \rowcolor[gray]{0.95} \multicolumn{3}{l}{\textit{NLP Tasks}} \\
    \quad classification & \cellcolor{highperf}98.21\% & \cellcolor{highperf}100.00\% \\
    \quad summarization & \cellcolor{highperf}94.83\% & \cellcolor{highperf}100.00\% \\
    \quad information\_extraction & \cellcolor{highperf}95.45\% & \cellcolor{highperf}95.45\% \\
    \quad question\_answering & \cellcolor{lowperf}85.07\% & \cellcolor{lowperf}86.51\% \\
    \quad not\_applicable & \cellcolor{lowperf}84.06\% & \cellcolor{lowperf}88.41\% \\
    \midrule
    \rowcolor[gray]{0.95} \multicolumn{3}{l}{\textit{Code Tasks}} \\
    \quad sql\_code\_generation & \cellcolor{highperf}95.83\% & \cellcolor{highperf}100.00\% \\
    \quad code\_generation & \cellcolor{lowperf}86.54\% & \cellcolor{lowperf}89.42\% \\
    \midrule
    \rowcolor[gray]{0.95} \multicolumn{3}{l}{\textit{Input Types}} \\
    \quad json & \cellcolor{highperf}96.88\% & \cellcolor{highperf}98.44\% \\
    \quad table & \cellcolor{goodperf}92.39\% & \cellcolor{highperf}97.83\% \\
    \quad knowledge\_base\_documents & \cellcolor{goodperf}92.02\% & \cellcolor{highperf}95.86\% \\
    \quad plain\_text & \cellcolor{midperf}89.24\% & \cellcolor{midperf}90.82\% \\
    \quad web\_search\_results & \cellcolor{lowperf}86.89\% & \cellcolor{midperf}90.16\% \\
    \midrule
    \rowcolor[gray]{0.95} \multicolumn{3}{l}{\textit{Output Requirements}} \\
    \quad long\_text & \cellcolor{goodperf}92.94\% & \cellcolor{midperf}90.98\% \\
    \quad markdown & \cellcolor{goodperf}92.21\% & \cellcolor{highperf}93.51\% \\
    \quad code\_snippet & \cellcolor{midperf}89.47\% & \cellcolor{goodperf}92.76\% \\
    \midrule
    \rowcolor[gray]{0.95} \multicolumn{3}{l}{\textit{Domains}} \\
    \quad sales\_marketing & \cellcolor{highperf}93.62\% & \cellcolor{highperf}100.00\% \\
    \quad data\_analytics & \cellcolor{goodperf}92.68\% & \cellcolor{highperf}98.78\% \\
    \quad mathematics & \cellcolor{midperf}88.95\% & \cellcolor{midperf}91.05\% \\
    \quad software\_development & \cellcolor{lowperf}86.01\% & \cellcolor{midperf}90.21\% \\
    \bottomrule
    \end{tabular}
\end{table}

\subsection{Taxonomy-Based Router}
\label{sec:taxonomy-router-results}

Table~\ref{tab:taxonomy-performance} summarizes the stratified performance of our taxonomy-based routing across 1,352 samples, revealing a bifurcation into two major groups:

\begin{itemize}[leftmargin=2em]
    \item \textbf{High-Performance Categories (43\% of tasks):}  
    Domains such as \texttt{classification} (98.21\% top-1), \texttt{arithmetic} (95.93\%), \texttt{summarization} (94.83\%), and \texttt{sql\_code\_generation} (95.83\%) consistently achieved top-1 accuracies above 92\%. These tasks tend to involve well-defined input-output requirements and more structured or constrained solution spaces, allowing taxonomy-based routing to quickly identify the best-suited model.

    \item \textbf{Variable-Performance Categories (57\% of tasks):}  
    Open-ended reasoning tasks (e.g., \texttt{causal\_reasoning} at 82.00\%) and knowledge-intensive domains like \texttt{question\_answering} (84.89\%) or multi-step procedures (88.61\%) show more variable performance with taxonomy-based routing alone. Such tasks often demand complex, multi-hop reasoning or specialized domain knowledge, making purely taxonomy-driven selection insufficient.
\end{itemize}

\paragraph{High-Performance Highlights}
Eight categories exceed 92\% top-1 accuracy, including \textbf{code/SQL generation} (up to 95.83\%), \textbf{arithmetic reasoning} (95.93\%), and \textbf{classification} (98.21\%). These tasks share structured output formats, formal constraints, or clearly bounded decision spaces, aligning well with the deterministic cues in our taxonomy framework. For instance, \texttt{format\_compliance} (92.16\%) benefits from explicit syntax rules, and \texttt{structured\_input} tasks (94.64\%) leverage well-defined schemas (e.g., JSON or tabular inputs).

\paragraph{Limitations in Lower-Performing Domains}
In contrast, 57\% of the dataset lies in categories where taxonomy-based routing alone is less effective. Here, tasks often involve \textbf{open-ended} or \textbf{multi-domain} reasoning (e.g., \texttt{causal\_reasoning}, 82.00\%), knowledge-intensive \texttt{question\_answering} (84.89\%), or tasks that require \textit{explanatory} outputs (e.g., \texttt{text\_with\_step\_by\_step\_explanation} at 83.10\%). Variance in performance is also higher ($\sigma^2=0.142$ vs.\ $0.037$ for high-performance categories), underlining the difficulty of matching ambiguous or cross-domain queries to a single optimal model via taxonomy alone.

These findings confirm that taxonomy-based routing can be both highly efficient and effective for approximately 43\% of tasks—those with clearer structure or well-defined domain cues—but struggles in more complex or open-ended settings. Consequently, this result motivates a \emph{learned router} or even a \emph{hybrid} approach, which augments taxonomy-based routing with data-driven selection signals for tasks where categorization alone is insufficient.

\subsection{Performance Across Routing Approaches}
\label{sec:learned-router-results}

Table~\ref{tab:router-comparison} presents our systematic evaluation of routing strategies, revealing distinct performance tiers and the compelling advantages of our hybrid approach.

\vspace{0.5em}
\noindent\textbf{Baseline Performance} 
Unsupervised methods demonstrate moderate effectiveness: hierarchical similarity matching using Voyage-3 embeddings (1024d)~\cite{voyageai2025voyage3large} with ScaNN (Scalable Nearest Neighbors) achieves 72.42\% accuracy, while direct fine-tuning of Llama-3.3-8B~\cite{grattafiori2024llama} with pairwise ranking loss for model selection reaches 71.35\%. Training our learned router exclusively on open-source data performs only marginally better (74.64\%). This consistent ceiling suggests fundamental limitations in generic data-driven approaches that lack domain-specific constraints.

\vspace{0.5em}
\noindent\textbf{Comparing Core Routing Approaches} 
Taxonomy-based routing—mapping queries to predefined categories with associated model preferences—achieves 77.92\% accuracy without cascading. The learned router architecture, which directly predicts model performance from query features, outperforms this baseline by 2.42 percentage points (80.34\%), highlighting the value of capturing nuanced query-model relationships.

\vspace{0.5em}
\noindent\textbf{Adding Cascading Mechanisms}
Both approaches demonstrate substantial gains when implementing cascading:
Taxonomy-based sees an increase accuracy to 86.29\% (+8.37 points) from a single cascade, while two cascades reach 88.05\% (+10.13 points). Our learned router achieves 89.21\% (+11.29 points) with a single cascade and 91.24\% (+13.32 points) with two cascades. These observed accuracy gains confirm the indispensable utility of structured fallback mechanisms in handling query diversity.

\vspace{0.5em}
\noindent\textbf{EmaFusion: Hybrid Routing with Cascades}
The integration of taxonomy-based domain knowledge with data-driven learning and two cascades achieves 94.25\% accuracy—a 16.33 percentage point improvement over the base taxonomy and 3.01 points beyond the learned router with two cascades. This performance jump stems from addressing complementary limitations:
\begin{itemize}
    \item Taxonomy-based methods provide structural inductive bias but lack flexibility for edge cases
    \item Learned systems offer adaptability but may struggle without appropriate domain constraints
    \item Cascading mechanisms enable systematic and often high-reward fallback when confidence is low
\end{itemize}

\vspace{0.5em}
\noindent\textbf{Empirical Validation}
These results strongly support the framework presented in Algorithm~\ref{alg:hybrid_routing}. The significant performance gap between baseline approaches ($\approx$72\%) and our hybrid method (94.25\%) demonstrates that effective model routing requires the integration of structured domain knowledge, data-driven learning, and cascade-based fallback mechanisms—each addressing different aspects of the routing challenge, with their integration yielding performance exceeding any single approach.

\definecolor{lowperf}{RGB}{255,235,235}
\definecolor{midperf}{RGB}{255,248,225}
\definecolor{highperf}{RGB}{240,255,240}
\definecolor{topperf}{RGB}{225,255,225}

\begin{table}[ht]
    \centering
    \caption{Performance comparison of routing approaches. The hybrid method combining taxonomy-based knowledge with learned routing and cascading mechanisms achieves the highest accuracy, demonstrating the complementary value of each component.}
    \label{tab:router-comparison}
    \begin{tabular}{lcc}
    \toprule
    \textbf{Approach} & \textbf{Accuracy} & \textbf{Improvement} \\
    \midrule
    \rowcolor[gray]{0.95} \multicolumn{3}{l}{\textit{Baseline Methods}} \\
    Clustering without Taxonomy & \cellcolor{lowperf}72.42\% & -5.50\% \\
    Fine-tuning for Top-3 Model Selection & \cellcolor{lowperf}71.35\% & -6.57\% \\
    Learned Router (Open Source Data only) & \cellcolor{lowperf}74.64\% & -3.28\% \\
    \midrule
    \rowcolor[gray]{0.95} \multicolumn{3}{l}{\textit{Taxonomy-Based Methods}} \\
    Taxonomy (no Cascading) & \cellcolor{lowperf}77.92\% & --- \\
    Taxonomy (Single Cascade) & \cellcolor{midperf}86.29\% & \cellcolor{midperf}+8.37\% \\
    Taxonomy (Two Cascades) & \cellcolor{midperf}88.05\% & \cellcolor{midperf}+10.13\% \\
    \midrule
    \rowcolor[gray]{0.95} \multicolumn{3}{l}{\textit{Learned Router Methods}} \\
    Learned Router (no Cascading) & \cellcolor{lowperf}80.34\% & \cellcolor{lowperf}+2.42\% \\
    Learned Router (Single Cascade) & \cellcolor{highperf}89.21\% & \cellcolor{highperf}+11.29\% \\
    Learned Router (Two Cascades) & \cellcolor{highperf}91.24\% & \cellcolor{highperf}+13.32\% \\
    \midrule
    \rowcolor[gray]{0.95} \multicolumn{3}{l}{\textit{Hybrid Approach}} \\
    \rowcolor[gray]{0.90} Taxonomy + Learned Router (Two Cascades) & \cellcolor{topperf}\textbf{94.25\%} & \cellcolor{topperf}\textbf{+16.33\%} \\
    \bottomrule
    \end{tabular}
\end{table}

\subsection{Performance Across Task Types}
Table~\ref{tab:learned-router-performance} presents the learned router performance across a strategically selected subset of task types from our complete evaluation dataset. These categories were chosen to highlight the model's behavior across the performance spectrum, illustrating key patterns observed in our taxonomy-based routing experiments. Overall, the learned router achieves a top-1 accuracy of 80.34\% and top-3 accuracy of 91.21\%.

\begin{table}[ht!]
    \centering
    \renewcommand{\arraystretch}{1.2}  %
    \setlength{\tabcolsep}{10pt}      %
    \caption{Learned Router Performance for High- vs. Variable-Performance Tasks. Tags indicate the broader category (Instruction Following, Knowledge Context Reasoning, Code Generation, Quantitative Analytical Reasoning) with lighter shades for Easy tasks and darker shades for Hard tasks.}
    \label{tab:learned-router-performance}
    \definecolor{ifeasy}{RGB}{65,105,225}      %
    \definecolor{ifhard}{RGB}{75,0,130}        %
    \definecolor{kceasy}{RGB}{255,140,0}       %
    \definecolor{kchard}{RGB}{178,34,34}       %
    \definecolor{qareasy}{RGB}{46,139,87}      %
    \definecolor{qarhard}{RGB}{0,100,0}        %
    \definecolor{code}{RGB}{138,43,226}        %
    
    \begin{tabular}{lccc}
    \toprule
    \textbf{Task (Tag)} & \textbf{Top-1} & \textbf{Top-3} & \textbf{Top-5} \\
    \midrule
    \multicolumn{4}{c}{\textit{High-Performance (Top-1 $\geq 90\%$)}} \\
    \midrule
    {\color{qareasy}[QAR-EASY]} GSM8K (Open-Source)       & 94.00\% & 98.00\% & 99.00\% \\
    {\color{ifeasy}[IF-EASY]} Multi-constraint Content Reformatting (Enterprise)        
                                           & 100.00\% & 100.00\% & 100.00\% \\
    {\color{ifeasy}[IF-EASY]} Conversational State Tracking (Enterprise)                
                                           & 100.00\% & 100.00\% & 100.00\% \\
    {\color{ifhard}[IF-HARD]} CX Ticket/Conversation Understanding (Enterprise)          
                                           & 100.00\% & 100.00\% & 100.00\% \\
    {\color{ifhard}[IF-HARD]} Proposal Document Refinement (Enterprise)                  
                                           & 92.00\%  & 96.00\%  & 98.00\% \\
    {\color{code}[CODE]} Technical Query Formulation (Enterprise)                 
                                           & 96.00\%  & 100.00\% & 100.00\% \\
    {\color{kchard}[KC-HARD]} RFP Generation (Enterprise)                                
                                           & 90.15\% & 92.63\%  & 100.00\% \\
    {\color{kceasy}[KC-EASY]} OPEN\_BENCHMARKS (Open-Source)                             
                                           & 95.74\%  & 97.87\%  & 100.00\% \\
    {\color{ifeasy}[IF-EASY]} Standardized Output Structuring (Enterprise)               
                                           & 95.56\%  & 97.78\%  & 97.78\% \\
    {\color{kceasy}[KC-EASY]} Hybrid Context Q\&A (Enterprise)                           
                                           & 93.02\%  & 95.35\%  & 95.35\% \\
    {\color{kchard}[KC-HARD]} Hierarchical Proposal Assistance (Enterprise)              
                                           & 92.86\%  & 92.86\%  & 97.62\% \\
    {\color{kceasy}[KC-EASY]} Context-aware Query Suggestion (Enterprise)                
                                           & 97.30\%  & 100.00\% & 100.00\% \\
    {\color{ifeasy}[IF-EASY]} Support Request Classification (Enterprise)                
                                           & 92.00\%  & 96.00\%  & 98.00\% \\
    {\color{ifeasy}[IF-EASY]} IFEval (Open-Source)                                        
                                           & 93.68\%  & 93.68\%  & 96.84\% \\
    {\color{ifeasy}[IF-EASY]} Executive Summary Generation (Enterprise)                  
                                           & 92.00\%  & 100.00\% & 100.00\% \\
    \midrule
    \multicolumn{4}{c}{\textit{Variable-Performance (Top-1 $< 90\%$)}} \\
    \midrule
    {\color{qareasy}[QAR-EASY]} ARC (Open-Source)                                           
                                           & 80.00\%  & 83.00\%  & 86.00\% \\
    {\color{qarhard}[QAR-HARD]} HLE (Open-Source)                                           
                                           & 68.37\%  & 66.33\%  & 77.55\% \\
    {\color{kchard}[KC-HARD]} Complex CX Resolution (Enterprise)                           
                                           & 78.00\%  & 86.00\%  & 90.00\% \\
    {\color{ifhard}[IF-HARD]} RFP Response Enhancement (Enterprise)                        
                                           & 80.00\%  & 80.00\%  & 85.00\% \\
    {\color{kchard}[KC-HARD]} Long-Context Understanding \& Generation (Enterprise)        
                                           & 83.78\%  & 91.89\%  & 94.59\% \\
    {\color{kceasy}[KC-EASY]} Search Relevance \& Reranking (Enterprise)                   
                                           & 71.43\%  & 95.24\%  & 95.24\% \\
    {\color{code}[CODE]} MBPP (Open-Source)                                         
                                           & 86.00\%  & 89.00\%  & 94.00\% \\
    {\color{kceasy}[KC-EASY]} MMLU (Open-Source)                                           
                                           & 86.57\%  & 89.55\%  & 94.03\% \\
    {\color{ifhard}[IF-HARD]} Tulu3 (Open-Source)                                          
                                           & 84.54\%  & 93.81\%  & 96.91\% \\
    \bottomrule
    \end{tabular}
\end{table}

\paragraph{Performance Stratification in Learned Router} Analysis of these results reveals a notable bifurcation in routing efficacy that aligns with our taxonomy-based findings (Section~\ref{sec:taxonomy-router-results}). The performance stratification is evident between well-structured tasks and those requiring open-ended reasoning, reflecting the fundamental pattern observed across our larger dataset (Section~\ref{sec:dataset}).

For well-structured tasks comprising approximately 43\% of our evaluation data, the router consistently demonstrates top-1 accuracy exceeding 88\%. These tasks—including mathematical reasoning (94.00\%), conversational state tracking (100.00\%), and content reformatting (98.00\%)—feature clear evaluation criteria, formal constraints, or bounded solution spaces. This aligns with previous findings in taxonomy-based routing approaches, where similar categories achieved top-1 accuracies above 92\%.

Conversely, for tasks requiring broader reasoning capabilities (approximately 57\% of our dataset), performance degrades considerably. Tasks like HLE (40.82\%), Complex CX Resolution (60.00\%), and ARC reasoning (63.00\%) present significant challenges for the router. These categories—characterized by open-ended reasoning, multi-domain knowledge integration, and explanatory outputs—present persistent challenges, mirroring limitations observed in taxonomy-based approaches.

The performance disparity across task types suggests that query-model matching operates on a spectrum of difficulty corresponding to task formalization. Tasks with well-defined success criteria facilitate more precise feature extraction and classification boundaries in the learned router. This observation extends prior work on meta-learning for model selection by demonstrating how task characteristics influence routing predictability. The substantial improvement from top-1 to top-3 accuracy (80.34\% → 91.21\%) further indicates that for many complex tasks, multiple models may perform comparably, making definitive single-model selection more challenging.

\paragraph{Cost Efficiency} Our analysis demonstrates an 82\% reduction in inference costs compared to an oracle approach (which would evaluate all models for every query), and a 68\% reduction compared to the average cost of randomly selecting models from our portfolio—all while maintaining comparable or superior performance for most query types. This efficiency emerges from the router's ability to distinguish between cases requiring high-capacity models and those where more efficient models suffice. The correlation between routing confidence and actual performance suggests the router has captured meaningful representations of model capabilities across the task space.

These observations motivated our investigation of hybrid approaches combining taxonomy-based routing with learned performance prediction, particularly for addressing the challenges in the "Variable-Performance" category.

\subsection{Hybrid: Taxonomy + Learned Routing}
\label{sec:hybrid-router-results}

\begin{figure}
    \centering
    \includegraphics[width=0.8\linewidth]{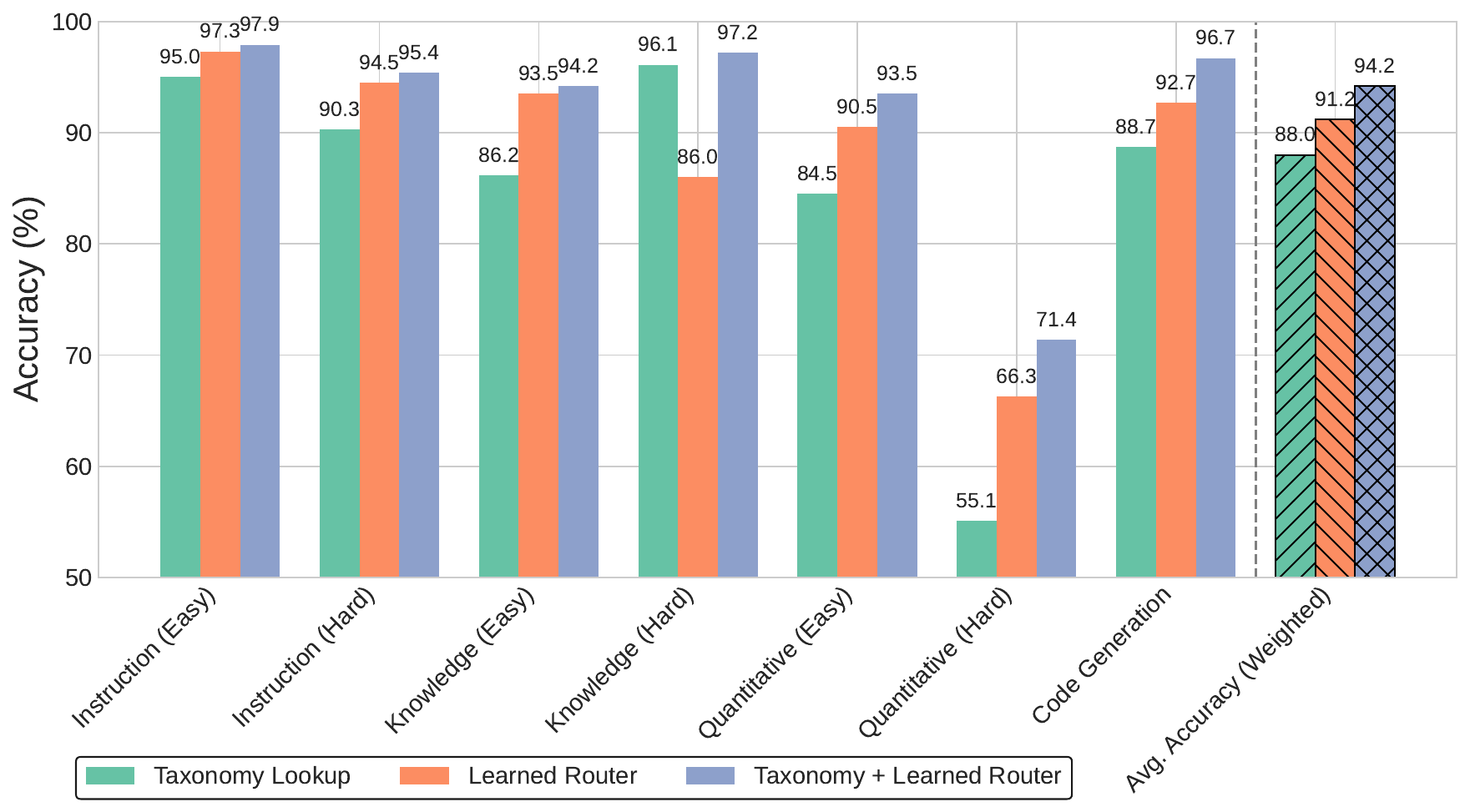}
    \caption{Comparison of hybrid approach (Taxonomy + Learned) with individual approaches of Taxonomy only and Learned only routing.}
    \label{fig:hybrid-category-comparison}
\end{figure}

Figure~\ref{fig:hybrid-category-comparison} compares three routing strategies---\emph{Taxonomy Lookup}, \emph{Predictor (Learned Router)}, and \emph{Taxonomy + Predictor} (our hybrid approach)---across seven representative categories. The hybrid method (green bars) consistently outperforms or matches the single-method baselines (blue and red bars), often by several percentage points:

\begin{itemize}[leftmargin=2em, itemsep=10pt, parsep=0pt, topsep=0pt]
    \item \textbf{Instruction (Easy / Hard).} 
    For straightforward instruction tasks, Taxonomy + Learned yields 97.9\% accuracy, a modest improvement over the Learned alone (97.3\%). Even for harder instruction-following queries, hybrid routing reaches 95.4\% , surpassing Learned’s 94.5\% by 0.9\%.

    \item \textbf{Knowledge (Easy / Hard).}
    The hybrid approach attains 94.2\% accuracy on simpler knowledge-based queries (a +0.8\% gain over 93.5\%), and a +1.1\% improvement (97.2\% vs. 96.1\%) on more challenging knowledge tasks. These gains highlight the benefit of combining taxonomic cues (e.g., domain labels) with a learned performance model.

    \item \textbf{Quantitative (Easy / Hard).}
    Simple numeric reasoning sees a +3.0\% jump (93.5\% vs.\ 90.5\%), while more difficult multi-step quantitative tasks improve even more (+5.1\%, from 66.3\% to 71.4\%). This suggests that taxonomy-based signals (e.g., \texttt{arithmetic}, \texttt{multi\_hop\_math}) help the learned router focus on models specialized in numeric reasoning, boosting overall accuracy.

    \item \textbf{Code Generation.}
    The hybrid method achieves 96.7\%, up from 92.7\% using the Learned alone, marking the largest improvement (+4.0\%). Code tasks often require domain-specific knowledge, consistent formatting, and robust error handling; combining taxonomy-based tagging with learned performance estimates is particularly advantageous.
\end{itemize}

Overall, the hybrid routing method leverages the strengths of both taxonomy-based model selection and learned performance prediction. Even in categories where one approach already excels, merging them typically yields small but meaningful boosts. In more complex or domain-specific tasks, these complementary signals can produce substantial gains, suggesting that hybrid routing offers a robust and versatile strategy for diverse query categories.

\subsection{EmaFusion: Hybrid + Cascading Router}

\begin{figure}[H]
    \centering
    \includegraphics[width=1.0\linewidth]{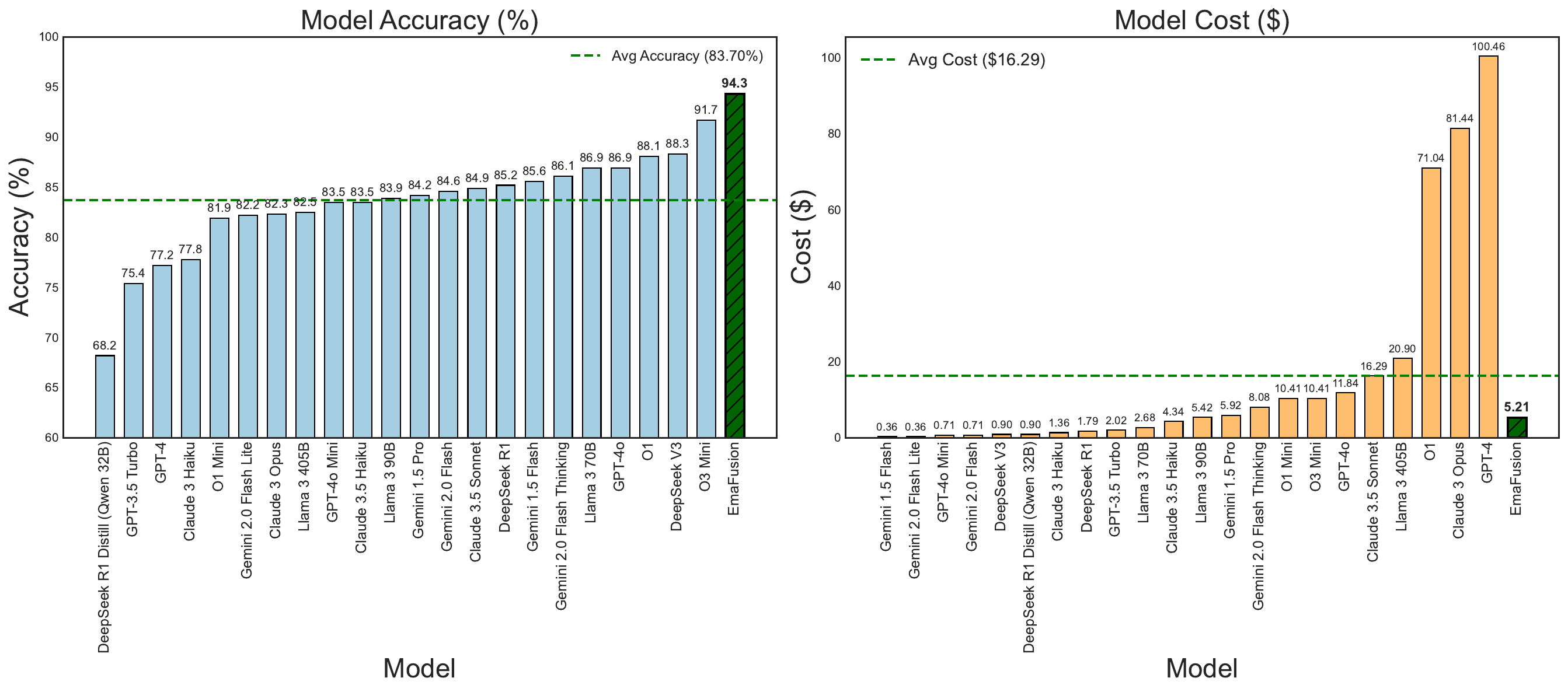}
    \caption{Comparison of cost and accuracy across our proposed \ema\ after cascading with other state-of-the-art models. 
    The average performance is denoted by the horizontal green line.}
    \label{fig:model-cost-accuracy}
\end{figure}

We compare the efficacy of the final part of the \ema\ pipeline by comparing the overall performance to the cost incurred. 

\paragraph{Cost Considerations.}
As shown in the right-hand chart of Figure~\ref{fig:model-cost-accuracy}, each bar represents the total monetary cost of running a particular model configuration on the same benchmark. The average cost across all models is \$17.00 (green dashed line). Notably, some premium-tier models (e.g., GPT-4, Claude~3~Opus) exceed \$70 in cost, while more budget-friendly or open-source solutions cost under \$1 but often exhibit significant performance drops. By contrast, \emph{EmaFusion} (highlighted in yellow) requires only \$5.38—less than one-third of the average, and an order of magnitude below the highest-tier commercial models. This low cost position underscores EmaFusion’s resource efficiency; it combines multiple specialized models and minimal overhead, thereby keeping inference expenditures modest relative to competing single-model solutions.

\paragraph{Accuracy Trade-Off.}
Meanwhile, the left-hand chart juxtaposes model accuracy rates, with 83.73\% as the overall mean (green dashed line). EmaFusion surpasses this average by over 11 percentage points, achieving 94.9\%. By comparison, models like Claude~3~Opus (92.7\% accuracy at \$81.44) or GPT-4 (88.7\% at \$100) illustrate the steep cost increases often required to achieve high accuracy. In other words, \emph{EmaFusion} strikes a particularly favorable \emph{cost-accuracy trade-off}—it matches or exceeds the performance of premium models without incurring premium costs. This advantage reflects its adaptive routing and fusion strategy, which smartly leverages the strengths of diverse specialized models to maximize accuracy while containing expenses. As a result, EmaFusion offers a compelling balance for users seeking top-tier performance with moderate resource requirements.

\section{Discussion}

\subsection{Evaluating Router Performance}
\label{sec:router-accuracy}

We evaluate our router from two complementary perspectives, \emph{sample-level} and \emph{model-level}. The former asks whether the router selects \emph{at least one} correct model per query; the latter analyzes how precisely and comprehensively the router includes each relevant model. Together, these analyses expose both the router’s effectiveness at “covering” good models and its capacity to exclude non-optimal ones.

\paragraph{Sample-Level Analysis.}
In the \emph{sample-level} view, we treat each query as a single instance and consider the router “correct” if any part of its selected set intersects the gold (top-performing) set. Formally:
\begin{itemize}[leftmargin=1.6em]
    \item \textbf{True Positive (TP)}: Prompt for which the router’s selection \emph{intersects} the gold set.
    \item \textbf{False Positive (FP)}: Prompt for which the router’s selection has \emph{no overlap} with the gold set.
    \item \textbf{Accuracy}: \(\text{TP} \,\big/\, \text{(Total Prompts)}\).
\end{itemize}

\begin{figure}[ht]
    \centering
    \begin{subfigure}[b]{0.48\textwidth}
        \centering
        \begin{minipage}[c][10cm][c]{\textwidth}  %
            \centering
            \includegraphics[width=\textwidth]{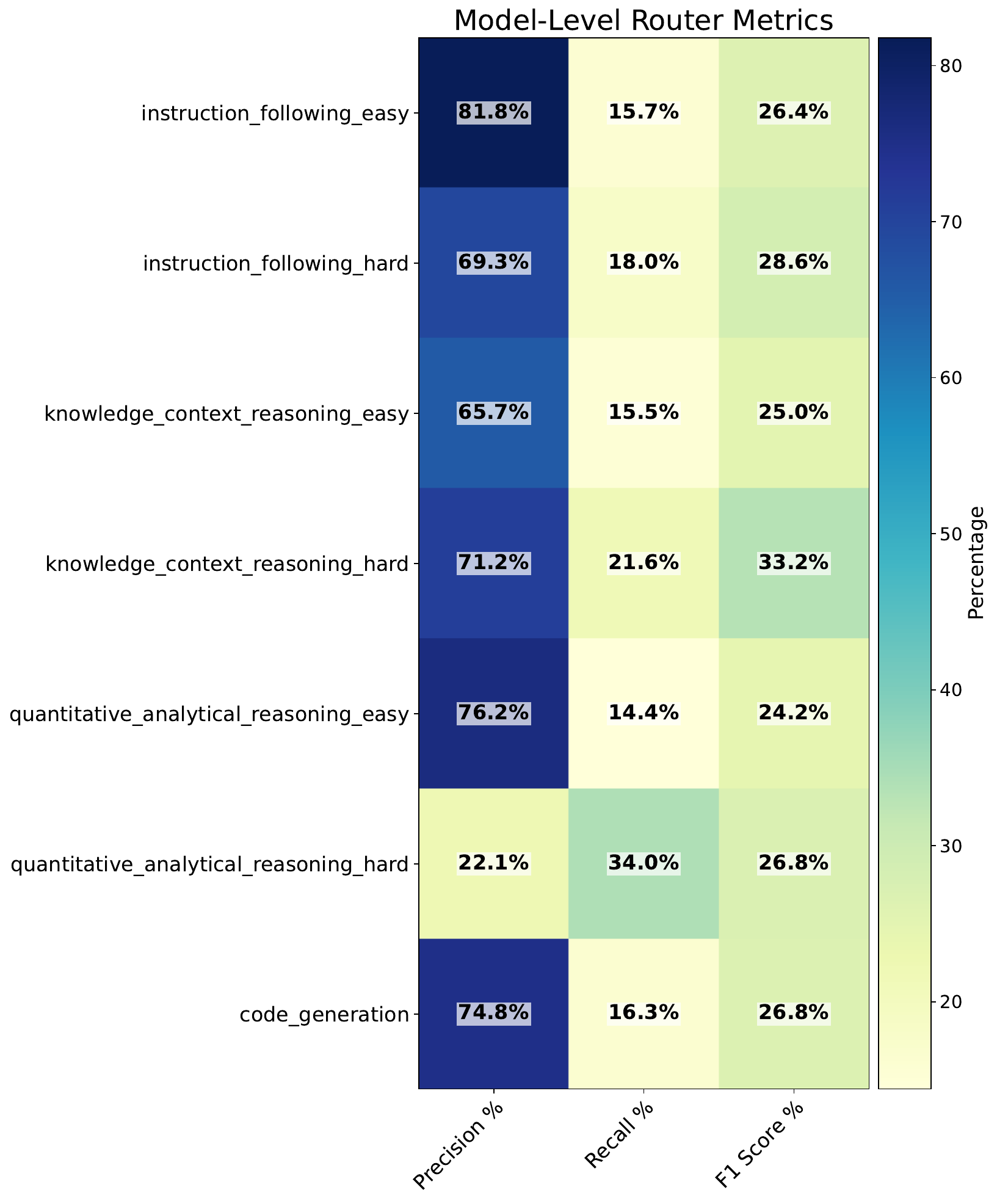}
        \end{minipage}
        \caption{}
        \label{fig:subfig1}
    \end{subfigure}
    \hfill
    \begin{subfigure}[b]{0.48\textwidth}
        \centering
        \begin{minipage}[c][10cm][c]{\textwidth}  %
            \centering
            \includegraphics[width=\textwidth]{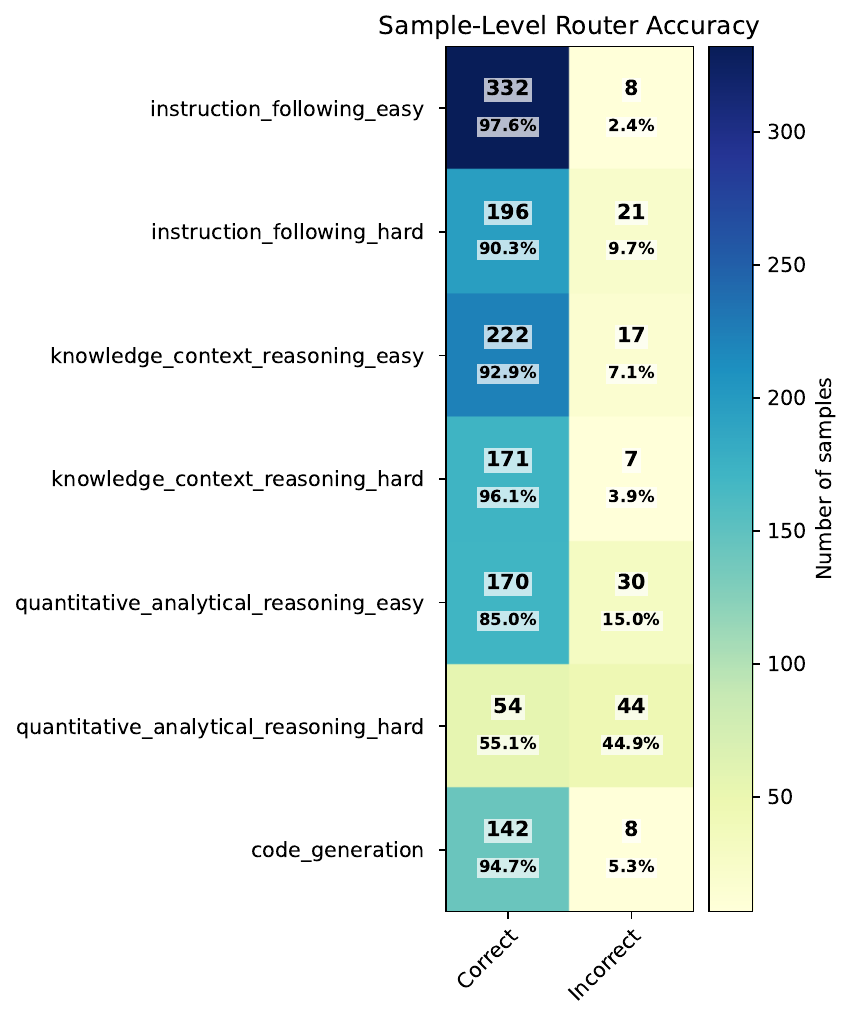}
        \end{minipage}
        \caption{}
        \label{fig:subfig2}
    \end{subfigure}
    \caption{Router Model vs Sample level accuracy comparison.}
    \label{fig:router-accuracy}
\end{figure}

This simple criterion yields direct insight into whether the router typically picks at least one strong model. Figure~\ref{fig:router-accuracy}(b) (right panel) shows that accuracy on tasks like \emph{instruction\_following\_easy} can exceed 95\%, while more complex domains such as \emph{quantitative\_analytical\_reasoning\_hard} hover near 55\%. Interestingly, some high-complexity tasks, for instance \emph{knowledge\_context\_reasoning\_hard}, still exhibit strong sample-level accuracy (~96\%), suggesting that robust domain cues can outweigh complexity when guiding the router to at least one suitable model. Overall, this analysis highlights the router’s reliability in covering plausible candidates, especially in domains that have clear patterns or format cues.

\paragraph{Model-Level Analysis}
Where sample-level metrics aggregate each query into a single pass/fail, our \emph{model-level} analysis drills down to every model decision. For each query--model pair, we note whether that model belongs to the gold set (i.e., it is among the top-performers for that query) and whether the router selected it. Specifically, we treat each model selection as a separate binary decision. For every model \(\{F_i\}\) and prompt:
\begin{itemize}[leftmargin=1.6em]
    \item \textbf{True Positive (TP)}: $F_i$ is in the set of top performing models and the router correctly includes $F_i$.
    \item \textbf{False Positive (FP)}: $F_i$ is not in the set of top performing models but the router incorrectly includes it.
    \item \textbf{True Negative (TN)}: $F_i$ is not in the set of top performing models and is correctly excluded.
    \item \textbf{False Negative (FN)}: $F_i$ is in the set of top performing models but is not selected by the router.
\end{itemize}

Summing true/false positives and negatives across all decisions within a task category produces micro-averaged \emph{precision}, \emph{recall}, and \emph{F1}. Figure~\ref{fig:router-accuracy}(a) (left panel) presents these metrics across the same categories. For instance, \emph{quantitative\_analytical\_reasoning\_hard} reveals low precision (22.1\%) but a moderate recall (34.0\%), indicating a tendency to over-select models; despite commonly including the correct one, the router also pulls in many extraneous models. By contrast, tasks like \emph{knowledge\_context\_reasoning\_hard} exhibit a higher precision (71.2\%) but a relatively limited recall (21.6\%), suggesting the router is more conservative—when it does select, it is usually correct, yet it sometimes misses relevant models for more ambiguous queries. 

This dual perspective (sample-level vs.\ model-level) clarifies the router’s behavior. A domain could show high sample-level accuracy if the router typically includes at least one strong model, yet still post lower precision or recall if it repeatedly picks additional suboptimal models or occasionally omits a key model. These trade-offs matter when balancing coverage (ensuring a good model is picked) against efficiency (reducing suboptimal model selections and executions).

\subsection{Number of Cascades impact on performance}
\begin{figure}[ht]
    \centering
    \begin{subfigure}[b]{0.48\textwidth}
        \centering
        \includegraphics[width=\textwidth]{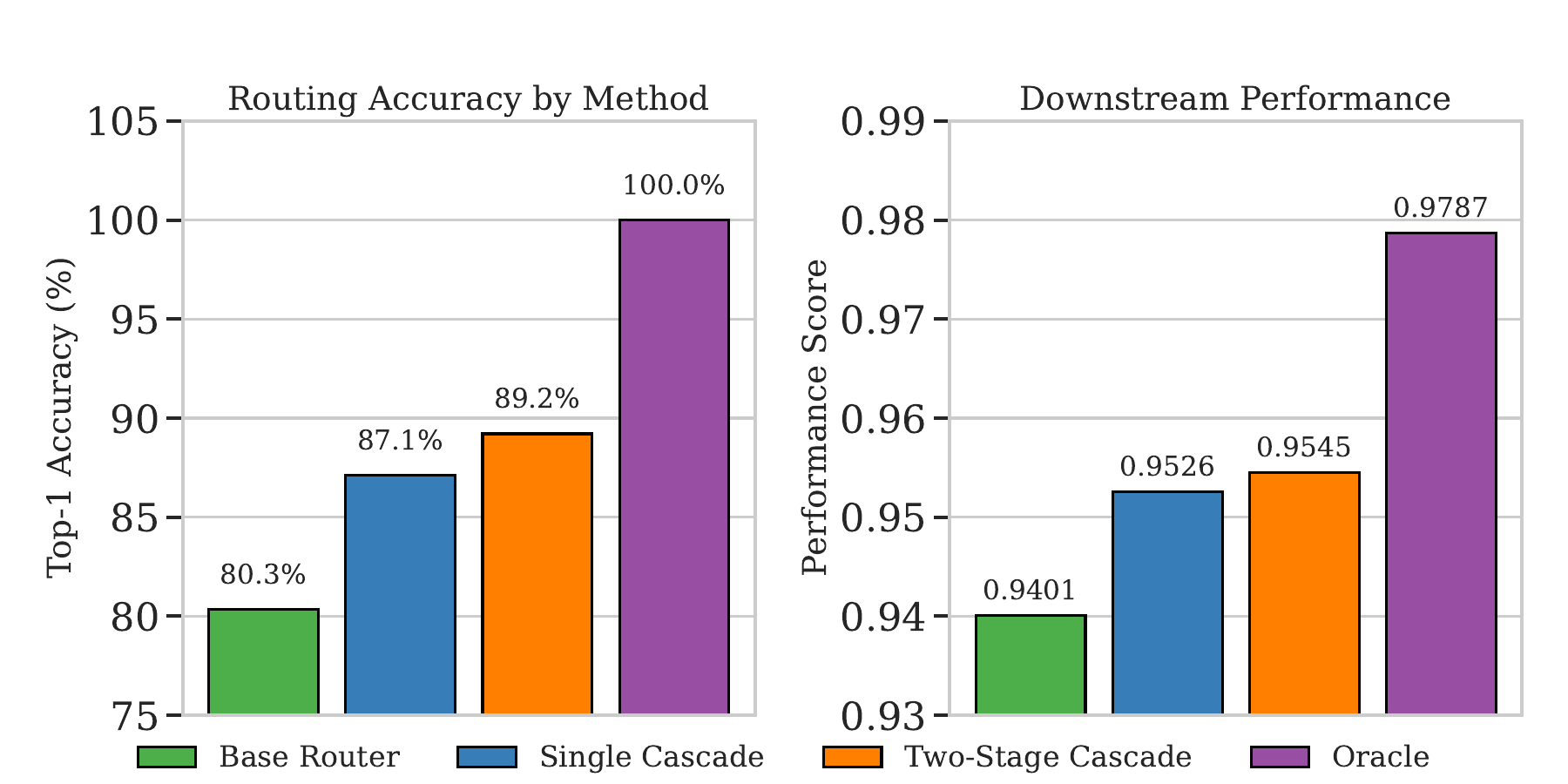}
        \caption{}
        \label{fig:subfig1}
    \end{subfigure}
    \hfill
    \begin{subfigure}[b]{0.48\textwidth}
        \centering
        \includegraphics[width=\textwidth]{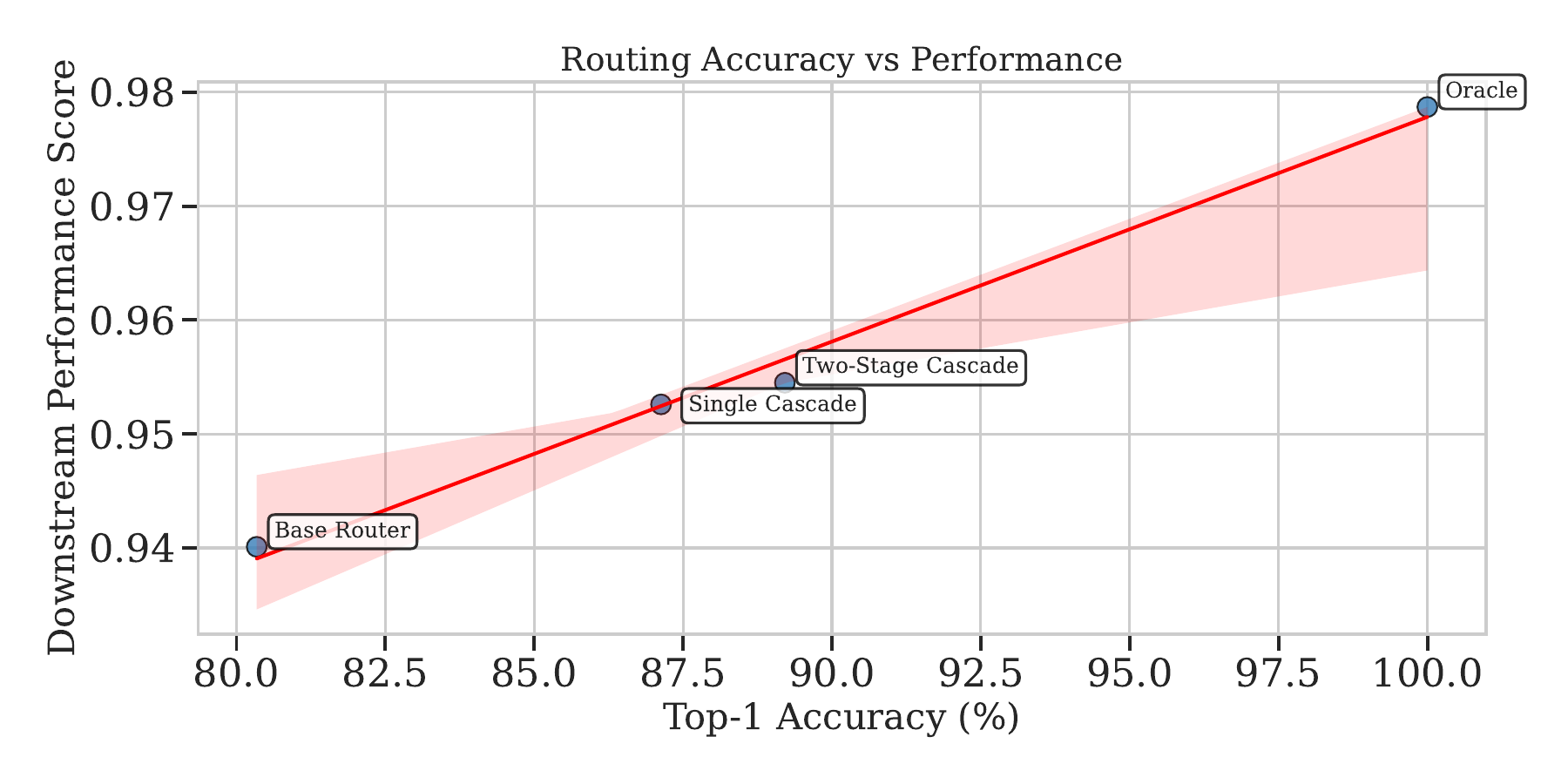}
        \caption{}
        \label{fig:subfig2}
    \end{subfigure}
    \caption{Visualizations of how cascading strategies affect routing precision and 
    overall performance. 
    Subfigure~(a) shows improvements in Top-1 Accuracy for Single and Two-Stage 
    Cascades (with optional termination). Subfigure~(b) illustrates the correlation 
    between accuracy and downstream performance.}
    \label{fig:cascades-performance}
\end{figure}

\paragraph{Impact of Single vs.\ Two-Stage Cascades.}
Figure~\ref{fig:cascades-performance}(a) compares the \emph{Base Router} (green) to one- and two-stage cascade variants (blue and orange), along with an \emph{Oracle} reference (purple). The \textbf{Base Router} alone achieves an 80.3\% top-1 routing accuracy, whereas \textbf{Single Cascade} raises this to 87.1\%. Introducing a \textbf{Two-Stage Cascade} results in a further gain to 89.2\%, but the improvement over a single cascade is relatively modest compared to the jump from the base router. Correspondingly, downstream performance scores show a similar pattern: moving from base to single cascade yields a clear increase (0.9401 to 0.9526), while a second cascade only slightly boosts that to 0.9545. 

\paragraph{Trade-Off and Recommended Stopping.}
As illustrated in Figure~\ref{fig:cascades-performance}(b), there is a strong correlation between routing accuracy (horizontal axis) and overall downstream performance (vertical axis). Although adding a second cascade does push both metrics higher, the marginal benefit tapers off beyond the single cascade. Hence, for most real-world scenarios, \emph{stopping after two cascades strikes a good balance} between improved accuracy/performance and additional inference cost or latency. After that point, further cascade stages yield diminishing returns, suggesting a practical sweet spot at one or two cascades.

\subsection{Distributional Analysis of Cascading Dynamics}

Our experimental findings reveal nuanced patterns in the distribution of cascading efficacy across the query space. The fact that 1.765 cascades are required, on average, to achieve top-5 level accuracy—and 3.57 for oracle-level performance—illuminates the diminishing returns inherent in extended cascading sequences.

Figure~\ref{fig:cascade_distribution} presents the probability mass function for the minimum number of cascades required to achieve optimal routing:

\begin{figure}[ht]
    \centering
    \begin{subfigure}[b]{0.48\textwidth}
        \centering
        \includegraphics[width=\textwidth]{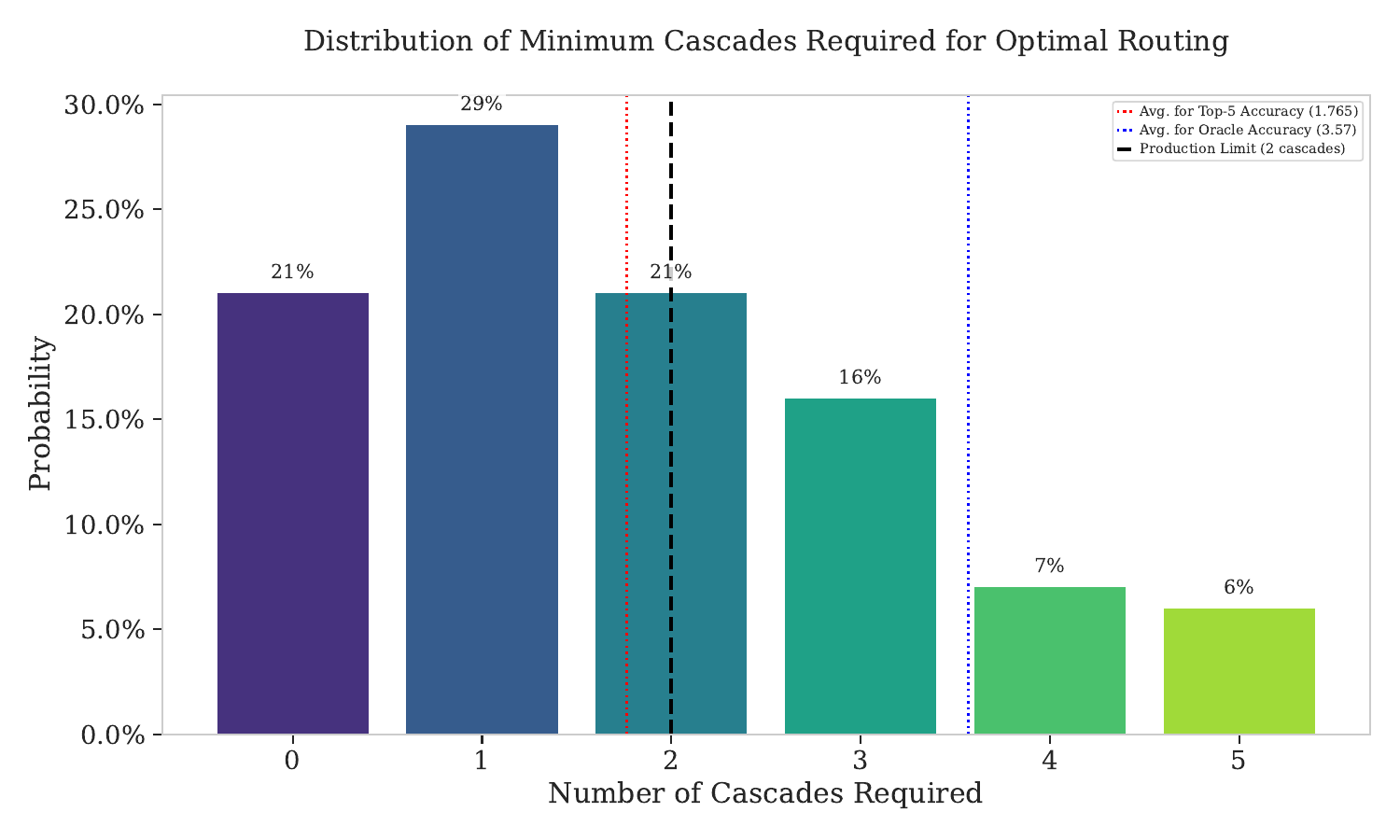}
        \caption{}
        \label{fig:subfig1}
    \end{subfigure}
    \hfill
    \begin{subfigure}[b]{0.48\textwidth}
        \centering
        \includegraphics[width=\textwidth]{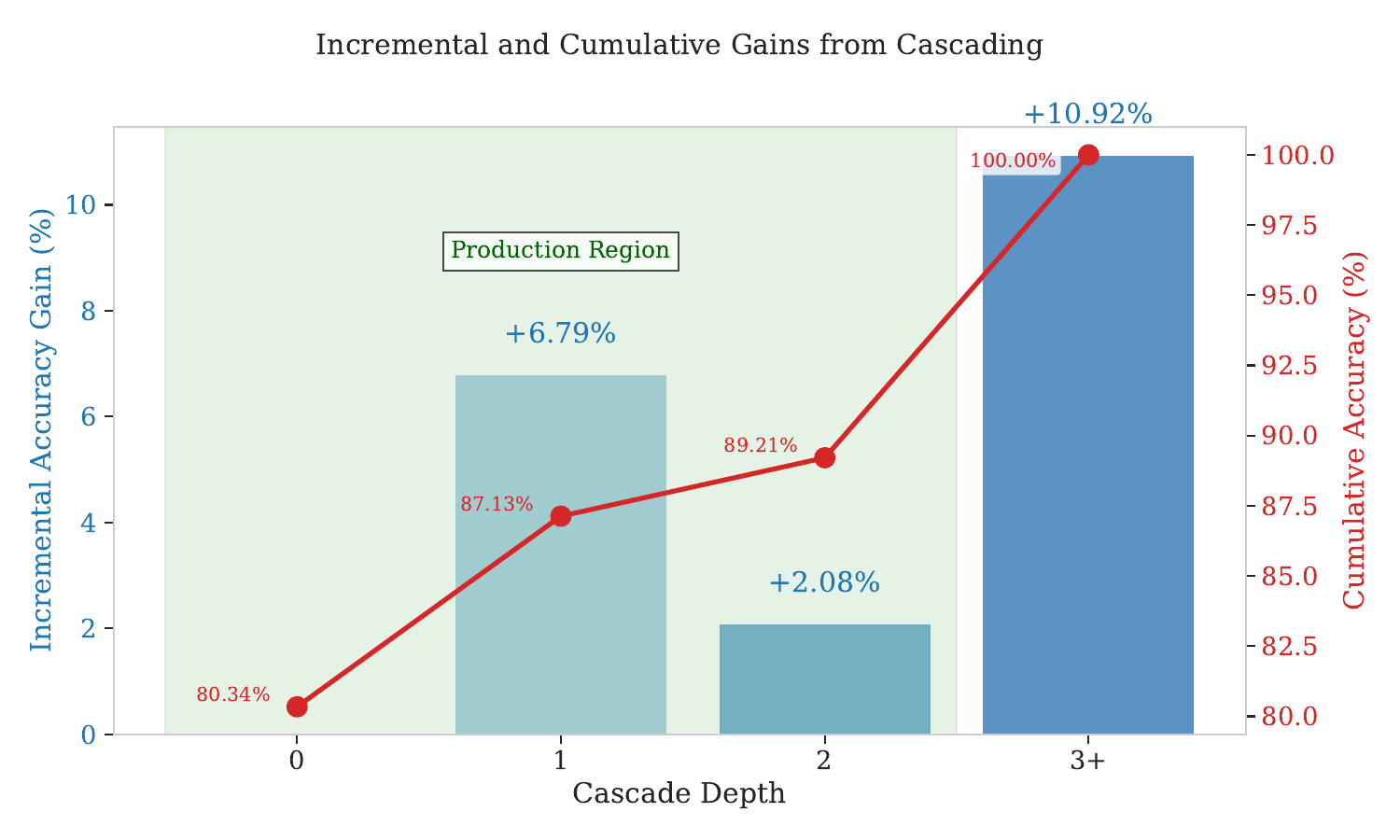}
        \caption{}
        \label{fig:subfig2}
    \end{subfigure}
    \caption{Distribution of Minimum Cascades Required for Optimal Routing}
    \label{fig:cascade_distribution}
\end{figure}

This distribution demonstrates a heavy left tail, with most corrections occurring within the first two cascades. The rapidly diminishing probability mass for $k > 2$ cascades empirically validates our production limitation of two cascades maximum—a constraint that captures 89.21\% of optimal routing decisions while maintaining operational efficiency.

The comparative analysis between single and two-stage cascading reveals an intriguing property: the incremental gain from the second cascade (2.08\% accuracy improvement) is substantially smaller than from the first (6.79\%). This sublinear improvement pattern suggests that the most egregious routing errors—those with the largest performance differentials—are typically corrected in the first cascade, with subsequent cascades addressing progressively more subtle misalignments.

\subsection{Economic Efficiency and Resource Utilization}

Beyond performance considerations, our cascading router demonstrates exceptional economic efficiency—a critical factor for production deployment at scale. Table~\ref{tab:economic_analysis} quantifies these advantages:

\begin{table}[h]
\centering
\caption{Economic Efficiency Analysis of Cascading Router}
\label{tab:economic_analysis}
\begin{tabular}{lrrl}
\toprule
\textbf{Cost Metric} & \textbf{Value} & \textbf{Comparison} & \textbf{Savings} \\
\midrule
Oracle evaluation cost & \$29.89 & --- & --- \\
Router prediction cost & \$5.21 & vs. Oracle & 82.57\% \\
GPT-4o baseline & \$11.84 & vs. GPT-4o & 56.00\% \\
Average model cost & \$16.29 & vs. Avg. Model & 68.00\% \\
\bottomrule
\end{tabular}
\end{table}

The economic advantages stem from two complementary factors:
\begin{enumerate}
    \item \textbf{Judicious Model Selection}: The router preferentially selects more economical models when their capabilities suffice for the query demands, reserving expensive models for genuinely complex queries.
    \item \textbf{Targeted Cascading}: By limiting cascades to situations where the reward model indicates suboptimal performance, the system minimizes unnecessary computation.
\end{enumerate}

This economic efficiency manifests as an 82.57\% cost reduction compared to the theoretical oracle approach—which would require evaluating all models to determine the optimal selection. When compared to the naïve approach of deploying a single premium model (GPT-4o) for all queries, our router delivers a 56.00\% cost reduction while maintaining comparable or superior performance for most query types.

\begin{figure}
    \centering
    \includegraphics[width=0.5\linewidth]{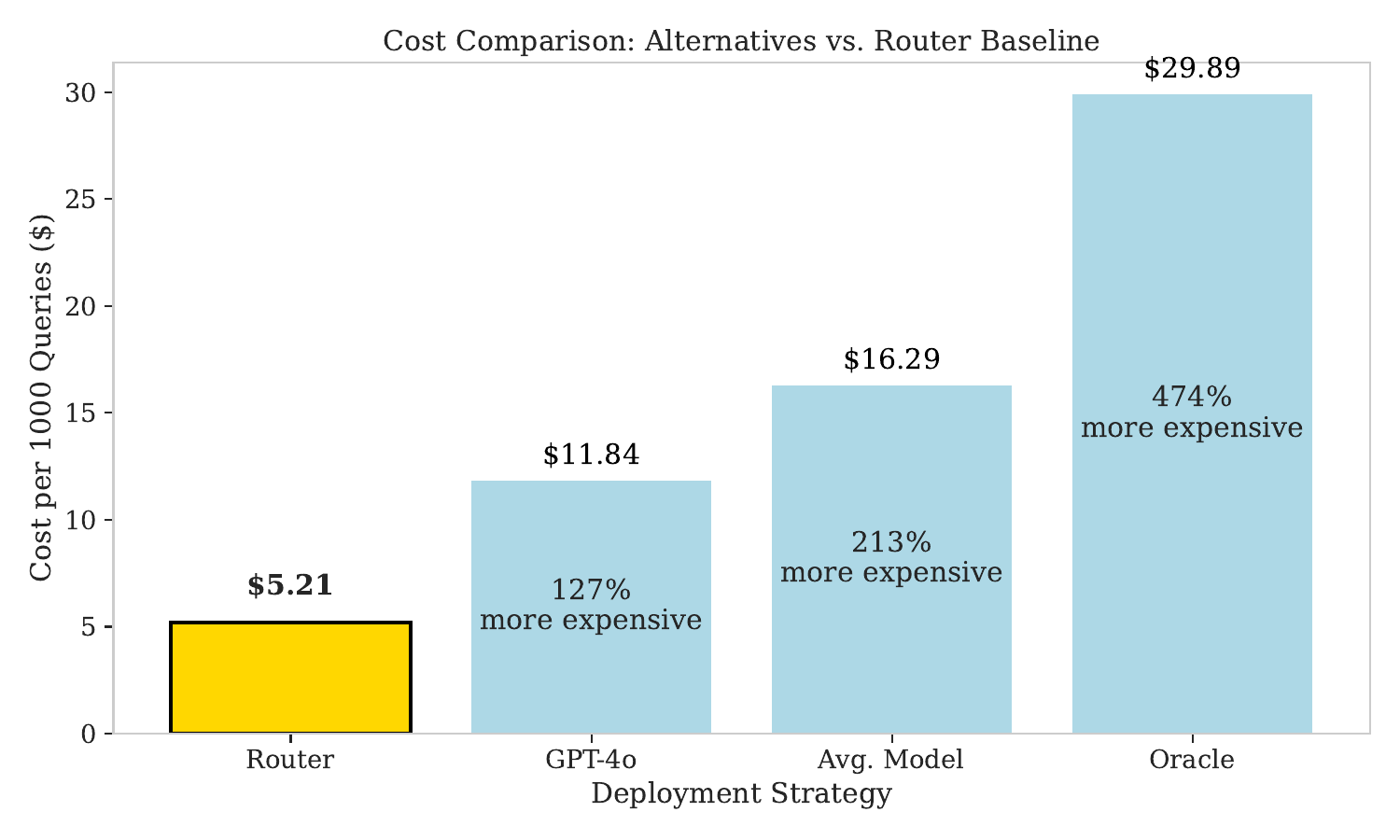}
    \caption{Economic Efficiency Analysis of Cascading Router}
    \label{fig:enter-label}
\end{figure}

\section{Conclusion}
In this work, we introduced \ema, a novel hybrid approach that synergistically integrates routing and fusion paradigms to enhance both accuracy and cost efficiency in the deployment of large language models (LLMs). Through intelligent routing based on model strengths and efficient fusion of their predictions, \ema\ achieves superior performance compared to existing methods. Empirical evaluations show that \ema\ outperforms the strongest single model (O3-mini) by 2.6 percentage points (94.3\% vs. 91.7\%) and exceeds the average model accuracy by over 10 points. At the same time, it provides substantial cost savings: \ema\ is roughly 3–4× cheaper than using an average model on every query, and nearly 20× cheaper than GPT‑4. Moreover, it supports flexible trade-offs, allowing users to fine-tune cost vs. accuracy based on the demands of each task.
In other words, \ema\ stands out as a practical, high-performing, and economically efficient solution for real-world applications that require strong accuracy without incurring excessive computational expenses.

\section{Future Work}
\label{sec:future-work}

\paragraph{Dynamic Integration of Newly Released Models.}
A key direction is to \emph{automate} the process of onboarding newly available foundation models—whether open-source or API-based—into the \ema\ framework. Currently, each new model must be manually benchmarked and incorporated into both the taxonomy suitability function \(\Phi(\cdot)\) and the learned router’s training pipeline. Future research could explore \textbf{online} learning procedure where \ema\ can continuously update its routing policy as new feedback on a model’s strengths is gathered. This means, for a new model, its released performance on various downstream tasks can be used to infer its likely strengths – and by running a \textit{few-shot calibration} on a small curated task set, its confidence can be assessed on each taxonomy category. 
Such adaptive approaches would facilitate \emph{real-time discovery} of whether a newly released model excels in particular tasks (e.g., code generation or multi-hop reasoning), allowing \ema\ to \emph{dynamically expand} its routing scheme.

\paragraph{Evolving Judge Models and Domain Experts.}
Since \ema\ relies heavily on judge-based signals, another major challenge lies in \emph{keeping these judges up to date}. Newly developed judge models or domain-specific verifiers might outperform existing ones, but integrating them can require extensive recalibration. One idea is to keep track of the mistakes that a specific judge is making. A self-training approach might work to improve such judges and involves few-shot updates: e.g., if the judges frequently mis-rank outputs in a certain domain, sampling multiple generations from the judge and training based on the corrective examples from that domain can help \cite{liu2024smart}. Another improvement is across the line of dynamically choosing the number of judges for a given task. An automated judge selection pipeline could decide, given the task type and available signals, which subset of judges (for example, general LLM, self-checking LLM, or domain-specific tool) to invoke for evaluation. Such a pipeline would streamline the CASCADE process: e.g., for a math problem, use an algebraic solver judge plus an LLM reasoning judge; for an open-ended question, use multiple LLM judges with diverse prompting (debate, critique, etc.) and so on. 

\section{Acknowledgment}

We thank all the team members at Ema Unlimited, Inc. for their valuable feedback and insightful discussions, which significantly improved this work. We especially thank Hemant Pugaliya, Narayanan Asuri Krishnan, Anshul Gupta, and Shobhit Saxena for their fruitful discussions. We also thank all our investors for their support and guidance.

\bibliography{ema_fusion}

\begin{thebibliography}{47}
\providecommand{\natexlab}[1]{#1}
\providecommand{\url}[1]{\texttt{#1}}
\expandafter\ifx\csname urlstyle\endcsname\relax
  \providecommand{\doi}[1]{doi: #1}\else
  \providecommand{\doi}{doi: \begingroup \urlstyle{rm}\Url}\fi

\bibitem[Achiam et~al.(2023)Achiam, Adler, Agarwal, Ahmad, Akkaya, Aleman, Almeida, Altenschmidt, Altman, Anadkat, et~al.]{achiam2023gpt}
Josh Achiam, Steven Adler, Sandhini Agarwal, Lama Ahmad, Ilge Akkaya, Florencia~Leoni Aleman, Diogo Almeida, Janko Altenschmidt, Sam Altman, Shyamal Anadkat, et~al.
\newblock Gpt-4 technical report.
\newblock \emph{arXiv preprint arXiv:2303.08774}, 2023.
\newblock URL \url{https://arxiv.org/abs/2303.08774}.

\bibitem[AI(2025)]{voyageai2025voyage3large}
Voyage AI.
\newblock Voyage-3-large: High-dimensional embedding models for advanced semantic representation.
\newblock \url{https://blog.voyageai.com/2025/01/07/voyage-3-large/}, 01 2025.
\newblock Accessed: 2024-10-30.

\bibitem[Aljundi et~al.(2017)Aljundi, Chakravarty, and Tuytelaars]{aljundi}
Rahaf Aljundi, Punarjay Chakravarty, and Tinne Tuytelaars.
\newblock Expert gate: Lifelong learning with a network of experts.
\newblock In \emph{2017 IEEE Conference on Computer Vision and Pattern Recognition (CVPR)}, pp.\  7120--7129, 2017.
\newblock \doi{10.1109/CVPR.2017.753}.
\newblock URL \url{https://ieeexplore.ieee.org/document/8100236}.

\bibitem[Anthropic(2024)]{anthropic2024claude3}
Anthropic.
\newblock Introducing the next generation of claude, Mar 2024.
\newblock URL \url{https://www.anthropic.com/news/claude-3-family}.
\newblock Accessed: 2025-03-07.

\bibitem[Austin et~al.(2021)Austin, Odena, Nye, Bosma, Michalewski, Dohan, Jiang, Cai, Terry, Le, et~al.]{mbpp}
Jacob Austin, Augustus Odena, Maxwell Nye, Maarten Bosma, Henryk Michalewski, David Dohan, Ellen Jiang, Carrie Cai, Michael Terry, Quoc Le, et~al.
\newblock Program synthesis with large language models.
\newblock \emph{arXiv preprint arXiv:2108.07732}, 2021.
\newblock URL \url{https://arxiv.org/abs/2108.07732}.

\bibitem[Chen et~al.(2024)Chen, Zhu, Dolan-Gavitt, Shafique, and Garg]{chen2024model}
Boyuan Chen, Mingzhi Zhu, Brendan Dolan-Gavitt, Muhammad Shafique, and Siddharth Garg.
\newblock Model cascading for code: Reducing inference costs with model cascading for llm based code generation.
\newblock \emph{arXiv preprint arXiv:2405.15842}, 2024.
\newblock URL \url{https://arxiv.org/abs/2405.15842v1}.

\bibitem[Chen et~al.(2023)Chen, Zaharia, and Zou]{chen2023frugalgpt}
Lingjiao Chen, Matei Zaharia, and James Zou.
\newblock Frugalgpt: How to use large language models while reducing cost and improving performance.
\newblock \emph{arXiv preprint arXiv:2305.05176}, 2023.
\newblock URL \url{https://arxiv.org/abs/2305.05176}.

\bibitem[Clark et~al.(2018)Clark, Cowhey, Etzioni, Khot, Sabharwal, Schoenick, and Tafjord]{allenai_arc}
Peter Clark, Isaac Cowhey, Oren Etzioni, Tushar Khot, Ashish Sabharwal, Carissa Schoenick, and Oyvind Tafjord.
\newblock Think you have solved question answering? try arc, the ai2 reasoning challenge.
\newblock \emph{arXiv:1803.05457v1}, 2018.
\newblock URL \url{https://arxiv.org/abs/1803.05457}.

\bibitem[Cobbe et~al.(2021)Cobbe, Kosaraju, Bavarian, Chen, Jun, Kaiser, Plappert, Tworek, Hilton, Nakano, Hesse, and Schulman]{cobbe2021gsm8k}
Karl Cobbe, Vineet Kosaraju, Mohammad Bavarian, Mark Chen, Heewoo Jun, Lukasz Kaiser, Matthias Plappert, Jerry Tworek, Jacob Hilton, Reiichiro Nakano, Christopher Hesse, and John Schulman.
\newblock Training verifiers to solve math word problems.
\newblock \emph{arXiv preprint arXiv:2110.14168}, 2021.
\newblock URL \url{https://arxiv.org/abs/2110.14168}.

\bibitem[Dekoninck et~al.(2024)Dekoninck, Baader, and Vechev]{dekoninck2024unified}
Jasper Dekoninck, Maximilian Baader, and Martin Vechev.
\newblock A unified approach to routing and cascading for llms.
\newblock \emph{arXiv preprint arXiv:2410.10347}, 2024.
\newblock URL \url{https://arxiv.org/abs/2410.10347}.

\bibitem[Du et~al.(2022)Du, Huang, Dai, Tong, Lepikhin, Xu, Krikun, Zhou, Yu, Firat, Zoph, Fedus, Bosma, Zhou, Wang, Wang, Webster, Pellat, Robinson, Meier-Hellstern, Duke, Dixon, Zhang, Le, Wu, Chen, and Cui]{du2022glam}
Nan Du, Yanping Huang, Andrew~M Dai, Simon Tong, Dmitry Lepikhin, Yuanzhong Xu, Maxim Krikun, Yanqi Zhou, Adams~Wei Yu, Orhan Firat, Barret Zoph, Liam Fedus, Maarten~P Bosma, Zongwei Zhou, Tao Wang, Emma Wang, Kellie Webster, Marie Pellat, Kevin Robinson, Kathleen Meier-Hellstern, Toju Duke, Lucas Dixon, Kun Zhang, Quoc Le, Yonghui Wu, Zhifeng Chen, and Claire Cui.
\newblock {GL}a{M}: Efficient scaling of language models with mixture-of-experts.
\newblock In \emph{Proceedings of the 39th International Conference on Machine Learning}, volume 162 of \emph{Proceedings of Machine Learning Research}, pp.\  5547--5569. PMLR, 17--23 Jul 2022.
\newblock URL \url{https://proceedings.mlr.press/v162/du22c.html}.

\bibitem[Du et~al.(2023)Du, Li, Torralba, Tenenbaum, and Mordatch]{du2023improving}
Yilun Du, Shuang Li, Antonio Torralba, Joshua~B Tenenbaum, and Igor Mordatch.
\newblock Improving factuality and reasoning in language models through multiagent debate.
\newblock \emph{arXiv preprint arXiv:2305.14325}, 2023.
\newblock URL \url{https://arxiv.org/abs/2305.14325}.

\bibitem[Fedus et~al.(2021)Fedus, Zoph, and Shazeer]{fedus2021switch}
William Fedus, Barret Zoph, and Noam Shazeer.
\newblock Switch transformers: Scaling to trillion parameter models with simple and efficient sparsity.
\newblock \emph{arXiv preprint cs.LG/2101.03961}, 2021.
\newblock URL \url{https://arxiv.org/abs/2101.03961}.

\bibitem[Feng et~al.(2025)Feng, Shen, and You]{feng2024graphrouter}
Tao Feng, Yanzhen Shen, and Jiaxuan You.
\newblock Graphrouter: A graph-based router for {LLM} selections.
\newblock In \emph{The Thirteenth International Conference on Learning Representations}, 2025.
\newblock URL \url{https://openreview.net/forum?id=eU39PDsZtT}.

\bibitem[Grattafiori et~al.(2024)Grattafiori, Dubey, Jauhri, Pandey, Kadian, Al-Dahle, Letman, Mathur, Schelten, Vaughan, et~al.]{grattafiori2024llama}
Aaron Grattafiori, Abhimanyu Dubey, Abhinav Jauhri, Abhinav Pandey, Abhishek Kadian, Ahmad Al-Dahle, Aiesha Letman, Akhil Mathur, Alan Schelten, Alex Vaughan, et~al.
\newblock The llama 3 herd of models.
\newblock \emph{arXiv preprint arXiv:2407.21783}, 2024.
\newblock URL \url{https://arxiv.org/abs/2407.21783}.

\bibitem[Guo et~al.(2025)Guo, Yang, Zhang, Song, Zhang, Xu, Zhu, Ma, Wang, Bi, et~al.]{guo2025deepseekR1}
Daya Guo, Dejian Yang, Haowei Zhang, Junxiao Song, Ruoyu Zhang, Runxin Xu, Qihao Zhu, Shirong Ma, Peiyi Wang, Xiao Bi, et~al.
\newblock Deepseek-r1: Incentivizing reasoning capability in llms via reinforcement learning.
\newblock \emph{arXiv preprint arXiv:2501.12948}, 2025.
\newblock URL \url{https://arxiv.org/abs/2501.12948}.

\bibitem[Hendrycks et~al.(2021)Hendrycks, Burns, Basart, Zou, Mazeika, Song, and Steinhardt]{mmlu}
Dan Hendrycks, Collin Burns, Steven Basart, Andy Zou, Mantas Mazeika, Dawn Song, and Jacob Steinhardt.
\newblock Measuring massive multitask language understanding.
\newblock In \emph{International Conference on Learning Representations}, 2021.
\newblock URL \url{https://openreview.net/forum?id=d7KBjmI3GmQ}.

\bibitem[Huang et~al.(2024)Huang, Feng, Li, Xiang, Wang, Liu, and Qin]{huang2024enabling}
Yichong Huang, Xiaocheng Feng, Baohang Li, Yang Xiang, Hui Wang, Ting Liu, and Bing Qin.
\newblock Ensemble learning for heterogeneous large language models with deep parallel collaboration.
\newblock In \emph{The Thirty-eighth Annual Conference on Neural Information Processing Systems}, 2024.
\newblock URL \url{https://openreview.net/forum?id=7arAADUK6D}.

\bibitem[Jiang et~al.(2023)Jiang, Ren, and Lin]{jiang2023llm}
Dongfu Jiang, Xiang Ren, and Bill~Yuchen Lin.
\newblock {LLM}-blender: Ensembling large language models with pairwise ranking and generative fusion.
\newblock In \emph{Proceedings of the 61st Annual Meeting of the Association for Computational Linguistics (Volume 1: Long Papers)}, pp.\  14165--14178. Association for Computational Linguistics, July 2023.
\newblock \doi{10.18653/v1/2023.acl-long.792}.
\newblock URL \url{https://aclanthology.org/2023.acl-long.792/}.

\bibitem[Kadavath et~al.(2022)Kadavath, Conerly, Askell, Henighan, Drain, Perez, Schiefer, Hatfield-Dodds, DasSarma, Tran-Johnson, et~al.]{kadavath2022language}
Saurav Kadavath, Tom Conerly, Amanda Askell, Tom Henighan, Dawn Drain, Ethan Perez, Nicholas Schiefer, Zac Hatfield-Dodds, Nova DasSarma, Eli Tran-Johnson, et~al.
\newblock Language models (mostly) know what they know.
\newblock \emph{arXiv preprint arXiv:2207.05221}, 2022.
\newblock URL \url{https://arxiv.org/abs/2207.05221}.

\bibitem[Kim et~al.(2024)Kim, Shin, Cho, Jang, Longpre, Lee, Yun, Shin, Kim, Thorne, and Seo]{kim2023prometheus}
Seungone Kim, Jamin Shin, Yejin Cho, Joel Jang, Shayne Longpre, Hwaran Lee, Sangdoo Yun, Seongjin Shin, Sungdong Kim, James Thorne, and Minjoon Seo.
\newblock Prometheus: Inducing fine-grained evaluation capability in language models.
\newblock In \emph{The Twelfth International Conference on Learning Representations}, 2024.
\newblock URL \url{https://openreview.net/forum?id=8euJaTveKw}.

\bibitem[Lambert et~al.(2024)Lambert, Morrison, Pyatkin, Huang, Ivison, Brahman, Miranda, Liu, Dziri, Lyu, et~al.]{tulu3-sft}
Nathan Lambert, Jacob Morrison, Valentina Pyatkin, Shengyi Huang, Hamish Ivison, Faeze Brahman, Lester James~V Miranda, Alisa Liu, Nouha Dziri, Shane Lyu, et~al.
\newblock T$\backslash$" ulu 3: Pushing frontiers in open language model post-training.
\newblock \emph{arXiv preprint arXiv:2411.15124}, 2024.
\newblock URL \url{https://arxiv.org/abs/2411.15124}.

\bibitem[Leviathan et~al.(2023)Leviathan, Kalman, and Matias]{leviathan2023fast}
Yaniv Leviathan, Matan Kalman, and Yossi Matias.
\newblock Fast inference from transformers via speculative decoding.
\newblock In Andreas Krause, Emma Brunskill, Kyunghyun Cho, Barbara Engelhardt, Sivan Sabato, and Jonathan Scarlett (eds.), \emph{Proceedings of the 40th International Conference on Machine Learning}, volume 202 of \emph{Proceedings of Machine Learning Research}, pp.\  19274--19286. PMLR, 23--29 Jul 2023.
\newblock URL \url{https://proceedings.mlr.press/v202/leviathan23a.html}.

\bibitem[Li et~al.(2024)Li, Zhang, Yu, Fu, and Ye]{li2024more}
Junyou Li, Qin Zhang, Yangbin Yu, Qiang Fu, and Deheng Ye.
\newblock More agents is all you need.
\newblock \emph{arXiv preprint arXiv:2402.05120}, 2024.
\newblock URL \url{https://arxiv.org/abs/2402.05120}.

\bibitem[Liu et~al.(2024{\natexlab{a}})Liu, Feng, Xue, Wang, Wu, Lu, Zhao, Deng, Zhang, Ruan, et~al.]{liu2024deepseek}
Aixin Liu, Bei Feng, Bing Xue, Bingxuan Wang, Bochao Wu, Chengda Lu, Chenggang Zhao, Chengqi Deng, Chenyu Zhang, Chong Ruan, et~al.
\newblock Deepseek-v3 technical report.
\newblock \emph{arXiv preprint arXiv:2412.19437}, 2024{\natexlab{a}}.
\newblock URL \url{https://arxiv.org/abs/2412.19437}.

\bibitem[Liu et~al.(2024{\natexlab{b}})Liu, Zeng, Liu, Yan, He, Wang, Yan, Liu, and Zhou]{liu2024skywork}
Chris~Yuhao Liu, Liang Zeng, Jiacai Liu, Rui Yan, Jujie He, Chaojie Wang, Shuicheng Yan, Yang Liu, and Yahui Zhou.
\newblock Skywork-reward: Bag of tricks for reward modeling in llms.
\newblock \emph{arXiv preprint arXiv:2410.18451}, 2024{\natexlab{b}}.
\newblock URL \url{https://arxiv.org/abs/2410.18451}.

\bibitem[Liu et~al.(2024{\natexlab{c}})Liu, Shridhar, Prajapat, Xia, and Sachan]{liu2024smart}
Rongxing Liu, Kumar Shridhar, Manish Prajapat, Patrick Xia, and Mrinmaya Sachan.
\newblock Smart: Self-learning meta-strategy agent for reasoning tasks.
\newblock \emph{arXiv preprint arXiv:2410.16128}, 2024{\natexlab{c}}.
\newblock URL \url{https://arxiv.org/abs/2410.16128}.

\bibitem[Loshchilov \& Hutter(2019)Loshchilov and Hutter]{loshchilov2017decoupled}
Ilya Loshchilov and Frank Hutter.
\newblock Decoupled weight decay regularization.
\newblock In \emph{International Conference on Learning Representations}, 2019.
\newblock URL \url{https://openreview.net/forum?id=Bkg6RiCqY7}.

\bibitem[Ong et~al.(2025)Ong, Almahairi, Wu, Chiang, Wu, Gonzalez, Kadous, and Stoica]{ong2024routellm}
Isaac Ong, Amjad Almahairi, Vincent Wu, Wei-Lin Chiang, Tianhao Wu, Joseph~E. Gonzalez, M~Waleed Kadous, and Ion Stoica.
\newblock Route{LLM}: Learning to route {LLM}s from preference data.
\newblock In \emph{The Thirteenth International Conference on Learning Representations}, 2025.
\newblock URL \url{https://openreview.net/forum?id=8sSqNntaMr}.

\bibitem[Phan et~al.(2025)Phan, Gatti, Han, Li, Hu, Zhang, Zhang, Shaaban, Ling, Shi, et~al.]{phan2025humanity}
Long Phan, Alice Gatti, Ziwen Han, Nathaniel Li, Josephina Hu, Hugh Zhang, Chen Bo~Calvin Zhang, Mohamed Shaaban, John Ling, Sean Shi, et~al.
\newblock Humanity's last exam.
\newblock \emph{arXiv preprint arXiv:2501.14249}, 2025.
\newblock URL \url{https://arxiv.org/abs/2501.14249}.

\bibitem[Shnitzer et~al.(2023)Shnitzer, Ou, Silva, Soule, Sun, Solomon, Thompson, and Yurochkin]{shnitzer2023large}
Tal Shnitzer, Anthony Ou, M{\'\i}rian Silva, Kate Soule, Yuekai Sun, Justin Solomon, Neil Thompson, and Mikhail Yurochkin.
\newblock Large language model routing with benchmark datasets.
\newblock \emph{arXiv preprint arXiv:2309.15789}, 2023.
\newblock URL \url{https://arxiv.org/abs/2309.15789}.

\bibitem[Shridhar et~al.(2022)Shridhar, Macina, El-Assady, Sinha, Kapur, and Sachan]{shridhar-etal-2022-automatic}
Kumar Shridhar, Jakub Macina, Mennatallah El-Assady, Tanmay Sinha, Manu Kapur, and Mrinmaya Sachan.
\newblock Automatic generation of socratic subquestions for teaching math word problems.
\newblock In \emph{Proceedings of the 2022 Conference on Empirical Methods in Natural Language Processing}, pp.\  4136--4149. Association for Computational Linguistics, December 2022.
\newblock \doi{10.18653/v1/2022.emnlp-main.277}.
\newblock URL \url{https://aclanthology.org/2022.emnlp-main.277/}.

\bibitem[Shridhar et~al.(2023)Shridhar, Jhamtani, Fang, Van~Durme, Eisner, and Xia]{shridhar2023screws}
Kumar Shridhar, Harsh Jhamtani, Hao Fang, Benjamin Van~Durme, Jason Eisner, and Patrick Xia.
\newblock Screws: A modular framework for reasoning with revisions.
\newblock \emph{arXiv preprint arXiv:2309.13075}, 2023.
\newblock URL \url{https://arxiv.org/abs/2309.13075}.

\bibitem[Shridhar et~al.(2024)Shridhar, Sinha, Cohen, Wang, Yu, Pasunuru, Sachan, Weston, and Celikyilmaz]{shridhar-etal-2024-art}
Kumar Shridhar, Koustuv Sinha, Andrew Cohen, Tianlu Wang, Ping Yu, Ramakanth Pasunuru, Mrinmaya Sachan, Jason Weston, and Asli Celikyilmaz.
\newblock The {ART} of {LLM} refinement: Ask, refine, and trust.
\newblock In \emph{Proceedings of the 2024 Conference of the North American Chapter of the Association for Computational Linguistics: Human Language Technologies (Volume 1: Long Papers)}, pp.\  5872--5883. Association for Computational Linguistics, June 2024.
\newblock \doi{10.18653/v1/2024.naacl-long.327}.
\newblock URL \url{https://aclanthology.org/2024.naacl-long.327/}.

\bibitem[Srivatsa et~al.(2024)Srivatsa, Maurya, and Kochmar]{srivatsa2024harnessing}
Kv~Aditya Srivatsa, Kaushal Maurya, and Ekaterina Kochmar.
\newblock Harnessing the power of multiple minds: Lessons learned from {LLM} routing.
\newblock In \emph{Proceedings of the Fifth Workshop on Insights from Negative Results in NLP}, pp.\  124--134. Association for Computational Linguistics, June 2024.
\newblock \doi{10.18653/v1/2024.insights-1.15}.
\newblock URL \url{https://aclanthology.org/2024.insights-1.15/}.

\bibitem[Team et~al.(2024)Team, Georgiev, Lei, Burnell, Bai, Gulati, Tanzer, Vincent, Pan, Wang, et~al.]{team2024gemini}
Gemini Team, Petko Georgiev, Ving~Ian Lei, Ryan Burnell, Libin Bai, Anmol Gulati, Garrett Tanzer, Damien Vincent, Zhufeng Pan, Shibo Wang, et~al.
\newblock Gemini 1.5: Unlocking multimodal understanding across millions of tokens of context.
\newblock \emph{arXiv preprint arXiv:2403.05530}, 2024.
\newblock URL \url{https://arxiv.org/abs/2403.05530}.

\bibitem[Thakur et~al.(2021)Thakur, Reimers, Daxenberger, and Gurevych]{thakur-2020-AugSBERT}
Nandan Thakur, Nils Reimers, Johannes Daxenberger, and Iryna Gurevych.
\newblock Augmented {SBERT}: Data augmentation method for improving bi-encoders for pairwise sentence scoring tasks.
\newblock In \emph{Proceedings of the 2021 Conference of the North American Chapter of the Association for Computational Linguistics: Human Language Technologies}, pp.\  296--310, Online, June 2021. Association for Computational Linguistics.
\newblock URL \url{https://www.aclweb.org/anthology/2021.naacl-main.28}.

\bibitem[Varangot-Reille et~al.(2025)Varangot-Reille, Bouvard, Gourru, Ciancone, Schaeffer, and Jacquenet]{varangot2025doing}
Clovis Varangot-Reille, Christophe Bouvard, Antoine Gourru, Mathieu Ciancone, Marion Schaeffer, and Fran{\c{c}}ois Jacquenet.
\newblock Doing more with less--implementing routing strategies in large language model-based systems: An extended survey.
\newblock \emph{arXiv preprint arXiv:2502.00409}, 2025.
\newblock URL \url{https://arxiv.org/abs/2502.00409}.

\bibitem[Wang et~al.(2024)Wang, Li, Shao, Xu, Dai, Li, Chen, Wu, and Sui]{wang2023math}
Peiyi Wang, Lei Li, Zhihong Shao, Runxin Xu, Damai Dai, Yifei Li, Deli Chen, Yu~Wu, and Zhifang Sui.
\newblock Math-shepherd: Verify and reinforce {LLM}s step-by-step without human annotations.
\newblock In \emph{Proceedings of the 62nd Annual Meeting of the Association for Computational Linguistics (Volume 1: Long Papers)}, pp.\  9426--9439. Association for Computational Linguistics, August 2024.
\newblock \doi{10.18653/v1/2024.acl-long.510}.
\newblock URL \url{https://aclanthology.org/2024.acl-long.510/}.

\bibitem[Wang et~al.(2023)Wang, Wei, Schuurmans, Le, Chi, Narang, Chowdhery, and Zhou]{wang2022self}
Xuezhi Wang, Jason Wei, Dale Schuurmans, Quoc~V Le, Ed~H. Chi, Sharan Narang, Aakanksha Chowdhery, and Denny Zhou.
\newblock Self-consistency improves chain of thought reasoning in language models.
\newblock In \emph{The Eleventh International Conference on Learning Representations}, 2023.
\newblock URL \url{https://openreview.net/forum?id=1PL1NIMMrw}.

\bibitem[Warner et~al.(2024)Warner, Chaffin, Clavi{\'e}, Weller, Hallstr{\"o}m, Taghadouini, Gallagher, Biswas, Ladhak, Aarsen, et~al.]{modernbert}
Benjamin Warner, Antoine Chaffin, Benjamin Clavi{\'e}, Orion Weller, Oskar Hallstr{\"o}m, Said Taghadouini, Alexis Gallagher, Raja Biswas, Faisal Ladhak, Tom Aarsen, et~al.
\newblock Smarter, better, faster, longer: A modern bidirectional encoder for fast, memory efficient, and long context finetuning and inference.
\newblock \emph{arXiv preprint arXiv:2412.13663}, 2024.
\newblock URL \url{https://arxiv.org/abs/2412.13663}.

\bibitem[Wei et~al.(2024)Wei, Karina, Chung, Jiao, Papay, Glaese, Schulman, and Fedus]{wei2024measuring}
Jason Wei, Nguyen Karina, Hyung~Won Chung, Yunxin~Joy Jiao, Spencer Papay, Amelia Glaese, John Schulman, and William Fedus.
\newblock Measuring short-form factuality in large language models.
\newblock \emph{arXiv preprint arXiv:2411.04368}, 2024.
\newblock URL \url{https://arxiv.org/abs/2411.04368}.

\bibitem[Xiao et~al.(2024)Xiao, Zhang, Wu, Xu, Wang, Han, Fu, Zhong, Zeng, Song, and Chen]{xiao2024chainofexperts}
Ziyang Xiao, Dongxiang Zhang, Yangjun Wu, Lilin Xu, Yuan~Jessica Wang, Xiongwei Han, Xiaojin Fu, Tao Zhong, Jia Zeng, Mingli Song, and Gang Chen.
\newblock Chain-of-experts: When {LLM}s meet complex operations research problems.
\newblock In \emph{The Twelfth International Conference on Learning Representations}, 2024.
\newblock URL \url{https://openreview.net/forum?id=HobyL1B9CZ}.

\bibitem[Yang et~al.(2023)Yang, Li, Zhou, Xiao, Fang, and Zhang]{Yang-medical-routing}
Han Yang, Mingchen Li, Huixue Zhou, Yongkang Xiao, Qian Fang, and Rui Zhang.
\newblock One llm is not enough: Harnessing the power of ensemble learning for medical question answering.
\newblock \emph{medRxiv}, 2023.
\newblock \doi{10.1101/2023.12.21.23300380}.
\newblock URL \url{https://www.medrxiv.org/content/early/2023/12/24/2023.12.21.23300380}.

\bibitem[Zhao et~al.(2024)Zhao, Jin, and Mao]{zhao2024eagle}
Zesen Zhao, Shuowei Jin, and Z~Morley Mao.
\newblock Eagle: Efficient training-free router for multi-llm inference.
\newblock \emph{arXiv preprint arXiv:2409.15518}, 2024.
\newblock URL \url{https://arxiv.org/abs/2409.15518}.

\bibitem[Zhou et~al.(2023{\natexlab{a}})Zhou, Sch{\"a}rli, Hou, Wei, Scales, Wang, Schuurmans, Cui, Bousquet, Le, and Chi]{zhou2022least}
Denny Zhou, Nathanael Sch{\"a}rli, Le~Hou, Jason Wei, Nathan Scales, Xuezhi Wang, Dale Schuurmans, Claire Cui, Olivier Bousquet, Quoc~V Le, and Ed~H. Chi.
\newblock Least-to-most prompting enables complex reasoning in large language models.
\newblock In \emph{The Eleventh International Conference on Learning Representations}, 2023{\natexlab{a}}.
\newblock URL \url{https://openreview.net/forum?id=WZH7099tgfM}.

\bibitem[Zhou et~al.(2023{\natexlab{b}})Zhou, Lu, Mishra, Brahma, Basu, Luan, Zhou, and Hou]{ifeval}
Jeffrey Zhou, Tianjian Lu, Swaroop Mishra, Siddhartha Brahma, Sujoy Basu, Yi~Luan, Denny Zhou, and Le~Hou.
\newblock Instruction-following evaluation for large language models, 2023{\natexlab{b}}.
\newblock URL \url{https://arxiv.org/abs/2311.07911}.

\end{thebibliography}
\bibliographystyle{iclr2025_conference}

\clearpage

\appendix
\section{Full Dataset Table}
\label{sec:full-dataset-table}

\vspace{0.5cm}

\subsection{Open-Source Datasets}
\label{sec:open-source-datasets}

\begin{table}[htbp]
\centering
\begin{tabular}{lrrrr}
\toprule
\cellcolor{gray!15}\textbf{Dataset} 
& \cellcolor{gray!15}\textbf{\# Train} & \cellcolor{gray!15}\textbf{Train \%} 
& \cellcolor{gray!15}\textbf{\# Test} & \cellcolor{gray!15}\textbf{Test \%} 
\\
\midrule
GSM8K \cite{cobbe2021gsm8k}
  & 1272 & 8.46\% & 100 & 6.38\% \\
\rowcolor{gray!5}
Tulu3 \cite{tulu3-sft}
  & 1004 & 6.68\% & 100 & 6.38\% \\
MMLU \cite{mmlu}
  &  982 & 6.53\% &  70 & 4.47\% \\
\rowcolor{gray!5}
IFEval \cite{ifeval}
  &  793 & 5.27\% & 100 & 6.38\% \\
MMLU Pro
  &  687 & 4.57\% &  49 & 3.13\% \\
\rowcolor{gray!5}
ARC \cite{allenai_arc}
  &  539 & 3.58\% & 100 & 6.38\% \\
Humanity Last Exam \cite{phan2025humanity}
  & --   &  --    & 100 & 6.38\% \\
\rowcolor{gray!5}
MBPP \cite{mbpp}
  & --   &  --    & 100 & 6.38\% \\
\bottomrule
\end{tabular}
\caption{\textbf{Open-Source Datasets.} This table shows the distribution of samples across public benchmark datasets in our training and test sets.}
\label{tab:open-source-datasets}
\end{table}

\subsection{Enterprise Tasks by Capability Domain}
\label{sec:enterprise-tasks}

\begingroup
\begin{longtable}{@{}lrrrr@{}}

\caption{\textbf{Enterprise Tasks by Capability Domain.} This table shows the distribution of proprietary enterprise tasks organized by functional capabilities. Tasks with test set samples are used for evaluation, while tasks without test samples were used only for training.} \label{tab:enterprise-tasks} \\

\toprule
\cellcolor{gray!15}\textbf{Category (Task)} 
& \cellcolor{gray!15}\textbf{\# Train} & \cellcolor{gray!15}\textbf{Train \%} 
& \cellcolor{gray!15}\textbf{\# Test} & \cellcolor{gray!15}\textbf{Test \%} 
\\
\midrule
\endhead

\multicolumn{5}{c}%
{{\tablename\ \thetable{} -- continued from previous page}} \\
\toprule
\cellcolor{gray!15}\textbf{Category (Task)} 
& \cellcolor{gray!15}\textbf{\# Train} & \cellcolor{gray!15}\textbf{Train \%} 
& \cellcolor{gray!15}\textbf{\# Test} & \cellcolor{gray!15}\textbf{Test \%} 
\\
\midrule
\endhead

\midrule \multicolumn{5}{c}{{Continued on next page}} \\ \midrule
\endfoot

\bottomrule
\endlastfoot

\multicolumn{5}{@{}l}{\cellcolor{gray!10}\textbf{\large Instruction Following (Easy)}} \\
Executive Summary Generation 
  & 231 & 1.54\% & 50 & 3.19\% \\
\rowcolor{gray!5}
Multi-constraint Content Reformatting 
  & 200 & 1.33\% & 50 & 3.19\% \\
Conversational State Tracking 
  & 200 & 1.33\% & 50 & 3.19\% \\
\rowcolor{gray!5}
Standardized Output Structuring 
  & 200 & 1.33\% & 50 & 3.19\% \\
Support Request Classification 
  & 200 & 1.33\% & 50 & 3.19\% \\
\rowcolor{gray!5}
Response Structure Templating 
  & 200 & 1.33\% & -- & -- \\
Instruction Generation \& Refinement 
  & 200 & 1.33\% & -- & -- \\
\rowcolor{gray!5}
Multi-category Classification 
  & 200 & 1.33\% & -- & -- \\
\addlinespace[0.5em]

\multicolumn{5}{@{}l}{\cellcolor{gray!10}\textbf{\large Instruction Following (Hard)}} \\
RFP Response Enhancement 
  & 493 & 3.28\% & 50 & 3.19\% \\
\rowcolor{gray!5}
CX Ticket/Conversation Understanding 
  & 200 & 1.33\% & 50 & 3.19\% \\
Proposal Document Refinement 
  & 200 & 1.33\% & 50 & 3.19\% \\
\rowcolor{gray!5}
Proposal Template Creation
  & 200 & 1.33\% & -- & -- \\
Contextual Query Rewriting
  & 200 & 1.33\% & -- & -- \\
\rowcolor{gray!5}
RFP Overview Summarization 
  & 200 & 1.33\% & -- & -- \\
System Prompt Engineering 
  & 200 & 1.33\% & -- & -- \\
\rowcolor{gray!5}
Document Structure Generation 
  & 200 & 1.33\% & -- & -- \\
RFP Section Classification 
  & 200 & 1.33\% & -- & -- \\
\rowcolor{gray!5}
Comprehensive Report Creation 
  & 200 & 1.33\% & -- & -- \\
\addlinespace[0.5em]

\multicolumn{5}{@{}l}{\cellcolor{gray!10}\textbf{\large Knowledge Context Reasoning (Easy)}} \\
Search Relevance \& Reranking 
  & 200 & 1.33\% & 48 & 3.06\% \\
\rowcolor{gray!5}
Context-aware Query Suggestion 
  & 200 & 1.33\% & 50 & 3.19\% \\
Hybrid Context Q\&A 
  & 200 & 1.33\% & 50 & 3.19\% \\
\rowcolor{gray!5}
Multi-modal Understanding \& Reranking 
  & 200 & 1.33\% & 51 & 3.25\% \\
Structured Data Extraction 
  & 200 & 1.33\% & -- & -- \\
\rowcolor{gray!5}
Entity Recognition \& Classification 
  & 200 & 1.33\% & -- & -- \\
Domain-specific Entity Extraction 
  & 200 & 1.33\% & -- & -- \\
\rowcolor{gray!5}
Context-based Resource Selection 
  & 200 & 1.33\% & -- & -- \\
Data Source Identification 
  & 200 & 1.33\% & -- & -- \\
\addlinespace[0.5em]

\multicolumn{5}{@{}l}{\cellcolor{gray!10}\textbf{\large Knowledge Context Reasoning (Hard)}} \\
Complex CX Resolution 
  & 200 & 1.33\% & 50 & 3.19\% \\
\rowcolor{gray!5}
Long-Context Understanding \& Generation 
  & 200 & 1.33\% & 49 & 3.13\% \\
Hierarchical Proposal Assistance 
  & 200 & 1.33\% & 50 & 3.19\% \\
\rowcolor{gray!5}
RFP Generation 
  & 200 & 1.33\% & 50 & 3.19\% \\
Knowledge-Augmented Responses 
  & 200 & 1.33\% & -- & -- \\
\rowcolor{gray!5}
Domain-Specific Information Extraction 
  & 200 & 1.33\% & -- & -- \\
Contextual Resource Retrieval 
  & 200 & 1.33\% & -- & -- \\
\rowcolor{gray!5}
Multi-source Response Synthesis 
  & 200 & 1.33\% & -- & -- \\
\addlinespace[0.5em]

\multicolumn{5}{@{}l}{\cellcolor{gray!10}\textbf{\large Quantitative Analytical Reasoning (Easy)}} \\
Conditional Data Filtering 
  & 200 & 1.33\% & -- & -- \\
\rowcolor{gray!5}
Data Feature Selection 
  & 200 & 1.33\% & -- & -- \\
Data Type Transformation 
  & 200 & 1.33\% & -- & -- \\
\addlinespace[0.5em]

\multicolumn{5}{@{}l}{\cellcolor{gray!10}\textbf{\large Quantitative Analytical Reasoning (Hard)}} \\
Data Visualization Generation 
  & 200 & 1.33\% & -- & -- \\
\rowcolor{gray!5}
Visualization Code Generation 
  & 200 & 1.33\% & -- & -- \\
Structured Data Generation 
  & 200 & 1.33\% & -- & -- \\
\rowcolor{gray!5}
Complex Query Decomposition
  & 200 & 1.33\% & -- & -- \\
Technical Solution Dimensioning 
  & -- & -- & -- & -- \\
\rowcolor{gray!5}
Technical Infrastructure Optimization 
  & -- & -- & -- & -- \\
\addlinespace[0.5em]

\multicolumn{5}{@{}l}{\cellcolor{gray!10}\textbf{\large Code Generation}} \\
Technical Query Formulation 
  & 240 & 1.60\% & 50 & 3.19\% \\
\rowcolor{gray!5}
SQL Debugging \& Repair
  & 200 & 1.33\% & -- & -- \\
Conversational Programming Assistant 
  & 200 & 1.33\% & -- & -- \\
\rowcolor{gray!5}
File System Analysis 
  & 200 & 1.33\% & -- & -- \\
Programming Interface Design 
  & 200 & 1.33\% & -- & -- \\
\end{longtable}
\endgroup

\subsection{Dataset Summary Statistics}
\label{sec:dataset-summary}

\begin{table}[htbp]
\centering
\begin{tabular}{lrr}
\toprule
\cellcolor{gray!15}\textbf{Measure} & \cellcolor{gray!15}\textbf{Training Set} & \cellcolor{gray!15}\textbf{Test Set} \\
\midrule
Total samples & 15,041 & 1,567 \\
\rowcolor{gray!5}
Open-source samples & 5,277 (35.1\%) & 619 (39.5\%) \\
Enterprise task samples & 9,764 (64.9\%) & 948 (60.5\%) \\
\rowcolor{gray!5}
Unique tasks/datasets & 51 & 25 \\
Category coverage & 7/7 (100\%) & 7/7 (100\%) \\
\bottomrule
\end{tabular}
\caption{\textbf{Dataset Summary.} Overview statistics highlighting the balance between open-source and enterprise tasks in our dataset.}
\label{tab:dataset-summary}
\end{table}

\clearpage

\section{Comprehensive Evaluation Methodology}
\label{app:eval-methodology}

This appendix details our structured evaluation framework for assessing model performance across multiple quality dimensions. Our approach addresses common challenges in LLM evaluation, including the need for consistency, explicit criteria, and nuanced assessment across different query types.

\subsection{Multi-dimensional Evaluation Framework}
\label{app:eval-framework}

We evaluate model responses using a comprehensive seven-dimension framework, each scored on a consistent scale:

\begin{itemize}
    \item \textbf{Instruction Following} (1-3): Assesses adherence to explicit and implicit task requirements.
    \item \textbf{Factual Correctness} (1-3): Measures accuracy of factual claims and absence of misinformation.
    \item \textbf{Reasoning Quality} (1-3): Evaluates logical consistency, appropriate inferences, and valid analytical approaches.
    \item \textbf{Completeness} (1-3): Determines whether all aspects of the query are adequately addressed.
    \item \textbf{Clarity \& Organization} (1-3): Assesses structure, coherence, and communicative effectiveness.
    \item \textbf{Relevance} (1-3): Measures focus and alignment with the user's informational needs.
    \item \textbf{Helpfulness} (Binary 0-1): Determines whether the response meaningfully assists the user in achieving their goal.
\end{itemize}

These dimensions are complemented by an \textbf{Overall Quality} score (1-5) that provides a more nuanced holistic assessment ranging from "poor" to "excellent."

\subsection{Query Type Classification and Dimension Weighting}
\label{app:query-classification}

Our framework classifies queries into five distinct types: factual, procedural, creative, analytical, and clarification. This classification is crucial as different dimensions receive varying weights depending on query type:

\begin{itemize}
    \item For \textbf{factual queries}, Factual Correctness and Reasoning Quality are prioritized, with factual errors heavily penalizing overall scores.
    \item For \textbf{procedural queries}, Instruction Following and Completeness receive greater emphasis.
    \item For \textbf{creative queries}, Relevance and Helpfulness are weighted more heavily, with greater tolerance for creative liberties.
\end{itemize}

This context-aware weighting ensures evaluations align with real-world user expectations for different query types.

\subsection{Ground Truth Assessment}
\label{app:ground-truth}

When a ground truth reference is available, our framework incorporates comparative assessment specifically for Factual Correctness and Reasoning Quality dimensions. The remaining dimensions continue to be evaluated based solely on how the response addresses the user's query and follows instructions, independent of ground truth.

This selective application of ground truth enables objective assessment of factual and reasoning aspects while preserving context-sensitive evaluation of other dimensions.

\subsection{Evaluation Prompt Template}
\label{app:eval-template}

Our evaluations are conducted using a structured prompt template that generates standardized JSON output. The template follows a modular design with the following components:

\begin{codehighlight}
\begin{lstlisting}[caption={Evaluation Prompt Template}, label={lst:eval-template}, style=templatecode]
You are an AI evaluation system designed to critically assess an AI assistant's responses to a task, based on specific instructions and user queries that describe the task. Your goal is to provide highly consistent, structured feedback across multiple dimensions.

<<EVALUATION_DIMENSIONS>>

<<QUERY_TYPE_CLASSIFICATION>>

<<DIMENSION_WEIGHTING>>

**SPECIAL CASE HANDLING**

For responses that:
- Contain technically correct information but potentially harmful advice: Score Helpfulness as 0 and note the issue
- Are excessively verbose but correct: May score lower on Clarity & Organization but not on Factual Correctness
- Are technically perfect but miss the intent of the query: Score lower on Relevance regardless of other qualities
- Refuse to answer inappropriate requests: Score highly on Instruction Following if the system instructions include such restrictions

**OUTPUT FORMAT REQUIREMENTS**

Your evaluation MUST follow this exact JSON structure:
{
    "query_type": "FACTUAL", 
    "dimensions": {
      "instruction_following": {
        "score": 2,
        "reasoning": "Specific reasoning for instruction following score"
      },
      "factual_correctness": {
        "score": 3,
        "reasoning": "Specific reasoning for factual correctness score"
      },
      "reasoning_quality": {
        "score": 2,
        "reasoning": "Specific reasoning for reasoning quality score"
      },
      "completeness": {
        "score": 3,
        "reasoning": "Specific reasoning for completeness score"
      },
      "clarity_organization": {
        "score": 2,
        "reasoning": "Specific reasoning for clarity and organization score"
      },
      "relevance": {
        "score": 3,
        "reasoning": "Specific reasoning for relevance score"
      },
      "helpfulness": {
        "score": 1,
        "reasoning": "Specific reasoning for helpfulness score"
      }
    },
    "overall_score": 3,
    "overall_reasoning": "Comprehensive explanation for your overall score"
}

When scoring, strictly adhere to these criteria. Do not invent intermediate scores or use ranges. Your evaluation must be highly consistent - the same quality of response should always receive the same scores.

<<GROUND TRUTH EXTENSION>>

\end{lstlisting}
\end{codehighlight}

In the sections that follow, we detail each component of this template.

\subsection{Components of our Judge Prompt}
\label{app:template-components}

\paragraph{Evaluation Dimensions}
\label{app:eval-dimensions}

The evaluation dimensions component defines the seven core assessment criteria and their scoring scales:

\begin{codehighlight}
\begin{lstlisting}[caption={Evaluation Dimensions Component}, label={lst:eval-dimensions}, style=maincode]
**Evaluation Dimensions:**

1. **Instruction Following** (1-3):
   - 1: FAILED - Ignores or misinterprets critical instructions, failing to meet key requirements
   - 2: PARTIAL - Follows some instructions but misses or incorrectly implements others
   - 3: SUCCESS - Correctly follows all important instructions as specified

2. **Factual Correctness** (1-3):
   - 1: INCORRECT - Contains significant factual errors that affect the overall validity
   - 2: PARTIALLY CORRECT - Contains minor factual errors or imprecisions that don't fundamentally undermine the response
   - 3: CORRECT - All factual claims are accurate with no notable errors

3. **Reasoning Quality** (1-3):
   - 1: FLAWED - Contains logical fallacies, significant reasoning errors, or invalid analytical approaches
   - 2: ACCEPTABLE - Shows generally sound reasoning with some minor logical gaps
   - 3: STRONG - Demonstrates excellent reasoning with valid logical steps and well-supported conclusions

4. **Completeness** (1-3):
   - 1: INCOMPLETE - Misses major aspects of the query, leaving important questions unanswered
   - 2: PARTIALLY COMPLETE - Addresses the main points but omits some relevant aspects
   - 3: COMPLETE - Thoroughly addresses all aspects of the query

5. **Clarity & Organization** (1-3):
   - 1: UNCLEAR - Poorly organized, confusing, or difficult to follow
   - 2: MODERATELY CLEAR - Generally understandable but with some organizational issues
   - 3: VERY CLEAR - Well-structured, coherent, and easy to understand

6. **Relevance** (1-3):
   - 1: IRRELEVANT - Mostly off-topic or fails to address the query
   - 2: PARTIALLY RELEVANT - Generally on-topic but contains irrelevant tangents
   - 3: RELEVANT - Focused entirely on addressing the query

7. **Helpfulness** (Binary):
   - 0: NOT HELPFUL - The response does not meaningfully help the user achieve their goal
   - 1: HELPFUL - The response provides genuine assistance toward the user's goal

**Overall Quality** (1-5): A more nuanced holistic assessment:
   - 1: POOR - Fundamentally flawed in multiple critical dimensions
   - 2: INADEQUATE - Major deficiencies that significantly impact usefulness
   - 3: ACCEPTABLE - Meets basic standards but has notable room for improvement
   - 4: GOOD - Strong across most dimensions with only minor weaknesses
   - 5: EXCELLENT - Exceptional quality across all relevant dimensions
\end{lstlisting}
\end{codehighlight}
\clearpage

\paragraph{Query Type Classification}
\label{app:eval-query-types}

The query type classification component helps categorize the nature of user queries for contextually appropriate evaluation:

\begin{codehighlight}
\begin{lstlisting}[caption={Query Type Classification Component}, label={lst:eval-query-classification}, style=maincode]
**QUERY TYPE CLASSIFICATION**

Before evaluating, classify the query into one of these types:
1. FACTUAL - Seeking objectively verifiable information
2. PROCEDURAL - Asking how to accomplish a specific task
3. CREATIVE - Requesting open-ended or generative content
4. ANALYTICAL - Requiring analysis, synthesis, or judgment
5. CLARIFICATION - Asking for explanation or elaboration
\end{lstlisting}
\end{codehighlight}

\paragraph{Dimension Weighting}
\label{app:eval-weighting}

The dimension weighting component establishes how different dimensions should be prioritized based on query type:

\begin{codehighlight}
\begin{lstlisting}[caption={Dimension Weighting Component}, label={lst:eval-weighting}, style=maincode]
**DIMENSION WEIGHTING**

The importance of each dimension depends on the query type:

For factual/informational queries:
- Factual Correctness and Reasoning Quality should be weighted most heavily
- A response with score 1 (INCORRECT) in Factual Correctness cannot receive an overall score higher than 2

For procedural/how-to queries:
- Instruction Following and Completeness should be weighted most heavily
- A response with score 1 (FAILED) in Instruction Following cannot receive an overall score higher than 2

For creative/open-ended queries:
- Relevance and Helpfulness should be weighted most heavily
- Lower Factual Correctness may be acceptable if the response is creative and helpful
\end{lstlisting}
\end{codehighlight}
\clearpage

\paragraph{Ground Truth Extension}
\label{app:ground-truth-extension}

When ground truth answers are available, we extend the base prompt template with additional instructions:

\begin{codehighlight}
\begin{lstlisting}[caption={Ground Truth Extension Component}, label={lst:ground-truth}, style=maincode]
**GROUND TRUTH ASSESSMENT INSTRUCTIONS**

You have been provided with a ground truth answer. Use this ONLY to assess:
- Factual Correctness
- Reasoning Quality

For all other dimensions, evaluate based SOLELY on how well the response addresses the user's query and follows instructions, WITHOUT reference to the ground truth.

When evaluating Factual Correctness:
- Score 1 (INCORRECT) if the response makes claims that directly contradict the ground truth on important matters
- Score 2 (PARTIALLY CORRECT) if the response has minor factual discrepancies with the ground truth
- Score 3 (CORRECT) if all claims align with the ground truth

When evaluating Reasoning Quality:
- Score 1 (FLAWED) if the reasoning process differs substantially from the ground truth in ways that lead to incorrect conclusions
- Score 2 (ACCEPTABLE) if the reasoning generally aligns with the ground truth but has minor logical gaps
- Score 3 (STRONG) if the reasoning closely matches the sound reasoning in the ground truth

Ground Truth Answer:
---------------------START OF GROUND TRUTH---------------------
{ground_truth}
---------------------END OF GROUND TRUTH---------------------
\end{lstlisting}
\end{codehighlight}

\subsection{Multi-Judge Evaluation System}
\label{app:multi-judge-system}

To ensure robust and reliable evaluations, we implement a multi-judge system that leverages multiple LLM evaluators, each with complementary capabilities.

\paragraph{Judge Model Selection}
\label{app:judge-selection}

Our multi-judge system employs:
\begin{itemize}
\item \texttt{o3-mini}: A fast, cost-efficient reasoning model optimized for STEM tasks such as science, math, and coding, along with an extended output capacity of up to 100K tokens. 
\item \texttt{claude-3-7-sonnet} (extended thinking): A hybrid reasoning model capable of quick responses or detailed, step-by-step reasoning. Its extended thinking mode excels in complex problem-solving, creative tasks, and coding workflows and has an output capacity up to 64K tokens.
\end{itemize}

Each (query, response) pair is judged by both models \emph{three times}, using slightly varied prompts (e.g., insertion of small textual noise) to reduce the effect of ephemeral sampling. We find that repeating each judge's evaluation yields significantly more stable scores.

\paragraph{Handling Aggregation Disagreements}
\label{app:disagreement-handling}

While the majority of multi-judge comparisons lie within an acceptable band of agreement (e.g., $\pm1$ for overall quality), more severe discrepancies trigger a fallback:
\begin{enumerate}
  \item We compute the difference $\Delta = |S_{\text{o3-mini}} - S_{\text{Claude}}|$.
  \item If $\Delta > 2$, the sample is flagged for an \emph{expert annotation pass}.
  \item If $1 < \Delta \leq 2$, we average dimension-level scores from the judge with lower standard deviation across repeated trials with those from the other judge's final score, effectively weighting more consistent judgments more heavily.
\end{enumerate}

This approach ensures that our evaluation system captures the nuances of response quality while maintaining consistency across different evaluation models.

For each evaluated model and query, we collect the structured JSON output, which is then aggregated to compute performance metrics across dimensions, query types, and task categories as described in Section \ref{sec:learned-router-results}.

In summary, while multi-LLM judging is not without challenges, it represents a tractable and effective approach for producing high-fidelity annotations in large-scale data pipelines, especially when combined with occasional human expert oversight. This comprehensive evaluation methodology provides the foundation for the performance data presented throughout our paper, enabling precise comparison of different routing strategies and model selection approaches.

\section{Appendix: Taxonomy-Based Routing Performance}
\label{sec:taxonomy-based-routing-performance-expanded}
This is an expanded view of the taxonomy-based routing discussed in section~\ref{sec:taxonomy-router-results}.

\definecolor{highperf}{RGB}{225,255,225}   
\definecolor{goodperf}{RGB}{245,255,240}   
\definecolor{midperf}{RGB}{255,248,225}    
\definecolor{lowperf}{RGB}{255,235,235}    

\begin{longtable}{p{7cm}cc}
    \caption{Performance Stratification Across Key Taxonomic Categories (1,352 samples). Performance values are color-coded by accuracy tier: dark green (>95\%), light green (92-95\%), yellow (85-92\%), and light red (<85\%).}
    \label{tab:taxonomy-performance2} \\
    
    \toprule
    \textbf{Category} & \textbf{Top-1 Acc.} & \textbf{Top-3 Acc.} \\
    \midrule
    \endfirsthead
    
    \multicolumn{3}{c}{\tablename\ \thetable{} -- continued from previous page} \\
    \toprule
    \textbf{Category} & \textbf{Top-1 Acc.} & \textbf{Top-3 Acc.} \\
    \midrule
    \endhead
    
    \midrule
    \multicolumn{3}{r}{\textit{Continued on next page}} \\
    \endfoot
    
    \bottomrule
    \endlastfoot
    
    \rowcolor[gray]{0.95} \multicolumn{3}{l}{\textit{Task Groups}} \\
    \quad quantitative\_analytical\_reasoning & \cellcolor{goodperf}93.89\% & \cellcolor{goodperf}94.66\% \\
    \quad instruction\_following: format\_policies & \cellcolor{midperf}91.43\% & \cellcolor{goodperf}92.87\% \\
    \quad instruction\_following: multi\_step\_procedure & \cellcolor{midperf}88.69\% & \cellcolor{midperf}90.72\% \\
    \quad knowledge\_context\_reasoning: RAG & \cellcolor{midperf}88.89\% & \cellcolor{midperf}88.89\% \\
    
    \rowcolor[gray]{0.95} \multicolumn{3}{l}{\textit{Reasoning Types}} \\
    \quad arithmetic & \cellcolor{highperf}95.93\% & \cellcolor{highperf}99.19\% \\
    \quad format\_compliance & \cellcolor{goodperf}92.16\% & \cellcolor{highperf}95.10\% \\
    \quad instruction\_analysis & \cellcolor{goodperf}93.75\% & \cellcolor{goodperf}93.06\% \\
    \quad instruction\_compliance & \cellcolor{midperf}90.08\% & \cellcolor{midperf}91.83\% \\
    \quad multi\_step\_reasoning & \cellcolor{midperf}88.38\% & \cellcolor{midperf}90.66\% \\
    \quad multi\_hop & \cellcolor{midperf}89.86\% & \cellcolor{midperf}88.02\% \\
    \quad mathematical\_reasoning & \cellcolor{midperf}85.09\% & \cellcolor{lowperf}83.33\% \\
    \quad causal\_reasoning & \cellcolor{lowperf}82.00\% & \cellcolor{midperf}86.00\% \\
    \quad domain\_specific\_reasoning & \cellcolor{lowperf}84.91\% & \cellcolor{midperf}86.79\% \\
    \quad deductive\_reasoning & \cellcolor{midperf}85.00\% & \cellcolor{midperf}90.00\% \\
    \quad constraint\_evaluation & \cellcolor{midperf}87.50\% & \cellcolor{goodperf}92.50\% \\
    
    \rowcolor[gray]{0.95} \multicolumn{3}{l}{\textit{NLP Tasks}} \\
    \quad classification & \cellcolor{highperf}98.21\% & \cellcolor{highperf}100.00\% \\
    \quad summarization & \cellcolor{goodperf}94.83\% & \cellcolor{highperf}100.00\% \\
    \quad text\_editing & \cellcolor{goodperf}94.23\% & \cellcolor{highperf}96.15\% \\
    \quad information\_extraction & \cellcolor{highperf}95.45\% & \cellcolor{highperf}95.45\% \\
    \quad natural\_language\_generation & \cellcolor{goodperf}92.20\% & \cellcolor{goodperf}92.20\% \\
    \quad question\_answering & \cellcolor{midperf}85.07\% & \cellcolor{midperf}86.51\% \\
    \quad not\_applicable & \cellcolor{lowperf}84.06\% & \cellcolor{midperf}88.41\% \\
    
    \rowcolor[gray]{0.95} \multicolumn{3}{l}{\textit{Code Tasks}} \\
    \quad sql\_debugging & \cellcolor{highperf}95.35\% & \cellcolor{highperf}100.00\% \\
    \quad sql\_code\_generation & \cellcolor{highperf}95.83\% & \cellcolor{highperf}100.00\% \\
    \quad code\_generation & \cellcolor{midperf}86.54\% & \cellcolor{midperf}89.42\% \\
    \quad no\_code\_tasks & \cellcolor{midperf}89.27\% & \cellcolor{midperf}91.06\% \\
    
    \rowcolor[gray]{0.95} \multicolumn{3}{l}{\textit{Input Types}} \\
    \quad json & \cellcolor{highperf}96.88\% & \cellcolor{highperf}98.44\% \\
    \quad markdown & \cellcolor{goodperf}92.98\% & \cellcolor{goodperf}93.39\% \\
    \quad table & \cellcolor{goodperf}92.39\% & \cellcolor{highperf}97.83\% \\
    \quad knowledge\_base\_documents & \cellcolor{goodperf}92.02\% & \cellcolor{highperf}95.86\% \\
    \quad plain\_text & \cellcolor{midperf}89.24\% & \cellcolor{midperf}90.82\% \\
    \quad web\_search\_results & \cellcolor{midperf}86.89\% & \cellcolor{midperf}90.16\% \\
    
    \rowcolor[gray]{0.95} \multicolumn{3}{l}{\textit{Output Requirements}} \\
    \quad list & \cellcolor{goodperf}94.03\% & \cellcolor{highperf}100.00\% \\
    \quad long\_text & \cellcolor{goodperf}92.94\% & \cellcolor{midperf}90.98\% \\
    \quad json & \cellcolor{goodperf}92.56\% & \cellcolor{highperf}96.74\% \\
    \quad markdown & \cellcolor{goodperf}92.21\% & \cellcolor{goodperf}93.51\% \\
    \quad text\_with\_structured\_formatting & \cellcolor{midperf}91.98\% & \cellcolor{goodperf}92.92\% \\
    \quad short\_text & \cellcolor{midperf}90.56\% & \cellcolor{goodperf}92.78\% \\
    \quad code\_snippet & \cellcolor{midperf}89.47\% & \cellcolor{goodperf}92.76\% \\
    \quad text\_with\_step\_by\_step\_explanation & \cellcolor{lowperf}83.10\% & \cellcolor{lowperf}84.76\% \\
    
    \rowcolor[gray]{0.95} \multicolumn{3}{l}{\textit{Domains}} \\
    \quad sales\_marketing & \cellcolor{goodperf}93.62\% & \cellcolor{highperf}100.00\% \\
    \quad customer\_support & \cellcolor{goodperf}92.93\% & \cellcolor{midperf}91.41\% \\
    \quad data\_analytics & \cellcolor{goodperf}92.68\% & \cellcolor{highperf}98.78\% \\
    \quad technical\_writing & \cellcolor{midperf}91.45\% & \cellcolor{goodperf}93.16\% \\
    \quad cybersecurity & \cellcolor{midperf}91.07\% & \cellcolor{highperf}98.21\% \\
    \quad project\_management & \cellcolor{midperf}90.00\% & \cellcolor{midperf}85.00\% \\
    \quad mathematics & \cellcolor{midperf}88.95\% & \cellcolor{midperf}91.05\% \\
    \quad social\_sciences\_humanities & \cellcolor{midperf}86.27\% & \cellcolor{midperf}88.24\% \\
    \quad software\_development & \cellcolor{midperf}86.01\% & \cellcolor{midperf}90.21\% \\
    \quad stem & \cellcolor{midperf}85.71\% & \cellcolor{lowperf}81.82\% \\
    \quad natural\_sciences & \cellcolor{lowperf}83.62\% & \cellcolor{lowperf}82.76\% \\
    \quad it\_infrastructure & \cellcolor{lowperf}82.43\% & \cellcolor{midperf}87.84\% \\
\end{longtable}
\clearpage

\section{Appendix: Qualitative Samples}

\begin{itemize}
\item \textbf{Example A: Simple Factual Query (Low Domain Complexity)}
  \begin{itemize}
    \item \textit{Prompt}: "What is the capital of France?"
    \item \underline{Process}:
      \begin{enumerate}
        \item Domain is classified as \emph{General Knowledge} (i.e., not specialized).
        \item Logit-based confidence $S_L$ is high (model is quite certain).
        \item Domain-specific verification $S_D$ is either not relevant or automatically passes (since it is a non-specialized domain).
        \item Self-reported confidence $S_S$ might also be high, e.g., the model says "I'm quite certain."
        \item Reward model $S_R$: Good coherence and clarity, so it returns a strong score.
        \item Weighted combination $S_{\text{combined}}$ is well above $\tau_{\text{borderline\_high}}$. 
        \item \textbf{Outcome}: The system \textbf{accepts} the answer immediately (no need to invoke $S_J$).
      \end{enumerate}
  \end{itemize}

\item \textbf{Example B: Complex Legal Query (Highly Specialized)}
  \begin{itemize}
    \item \textit{Prompt}: "Under the 2022 Data Protection Act, which articles apply to cross-border data transfers in sub-Saharan Africa?"
    \item \underline{Process}:
      \begin{enumerate}
        \item Domain is classified as \emph{Legal Specialized}.
        \item $S_L$ might be moderate (the model is uncertain).
        \item Domain-specific verification $S_D$ is triggered (system cross-checks if references align with known legal statutes).
        \item If $S_D$ is below $\tau_{\text{domain}}$, the system \textbf{defers} or escalates to a specialized model (or next cascade model).
        \item Suppose, after re-routing or using a specialized model, $S_D \approx 0.75$ but $S_L$ is only 0.5, $S_S$ is 0.3 (model admits partial uncertainty).
        \item Weighted sum $S_{\text{combined}}$ falls into the borderline range $[\tau_{\text{borderline\_low}}, \tau_{\text{borderline\_high}}]$.
        \item System invokes $S_J$, the LLM judge. The judge examines the response for completeness and correctness.
        \item If $S_J < 0.5$, the system defers further. Otherwise, it accepts the specialized model's answer.
      \end{enumerate}
  \end{itemize}

\item \textbf{Example C: Medical QA with Overconfidence Issue}
  \begin{itemize}
    \item \textit{Prompt}: "What is the optimal insulin dosage for a Type 1 diabetes patient weighing 70 kg?"
    \item \underline{Process}:
      \begin{enumerate}
        \item Domain is classified as \emph{Medical Specialized}.
        \item $S_L$ (logit confidence) might be misleadingly high because the model is generically certain in numerical answers.
        \item $S_D$ (domain verification) fails, e.g., the system sees conflicting or contradictory medical references.
        \item The model also has a self-reported confidence $S_S$ that is moderate, but it references disclaimers about not being a medical professional.
        \item Weighted sum $S_{\text{combined}}$ is borderline or possibly below acceptance threshold.
        \item The system defers to LLM judge $S_J$ or escalates to a more specialized medical model. 
        \item Possibly the final outcome is \textbf{"No confident solution"} or a fallback prompt clarifying disclaimers.
      \end{enumerate}
  \end{itemize}

\item \textbf{Example D: Multi-Domain Cascade with Contract Generation}
  \begin{itemize}
    \item \textit{Prompt}: 
      \texttt{"Draft a legal contract concerning a new medical device, referencing ISO-13485 and HIPAA regulations, with bullet points for risk management."}
    \item \underline{Process}:
      \begin{enumerate}
        \item \textbf{Domain Classification:} The system detects \emph{legal + medical} keywords and routes the request first to a \emph{Tier-2 specialized legal-LLM} with partial medical knowledge.
        \item \textbf{Pre-Gen Gating:} Resource constraints allow proceeding. The user’s monthly budget is still sufficient.
        \item \textbf{During Generation (Chunk-Level Checks):}
          \begin{itemize}
            \item \emph{Regulatory Overlap Check:} The system identifies references to HIPAA and performs a partial $S_D$ to verify consistency with HIPAA sections. No mismatches are found.
            \item \emph{Incoherence Check:} No disclaimers or repetitive loops detected; generation continues.
          \end{itemize}
        \item \textbf{Post-Generation (CASCADE Signals):}
          \begin{itemize}
            \item $S_L$ is moderately high (0.8).
            \item $S_S$ (self-reported confidence) is 0.6, indicating some uncertainty about the risk management structure.
            \item $S_D$ is 0.75, as domain checks find mostly valid references.
            \item $S_R$ is 0.85 (reward model favors the style).
            \item Weighted sum $S_{\text{combined}} \approx 0.75$, placing it just inside the borderline range [0.7, 0.8].
          \end{itemize}
          The system invokes the LLM judge $S_J$, which returns 0.4 (some ISO standards references are incomplete). Since $S_J < 0.5$, the final decision is \emph{defer}.
        \item \textbf{Escalation:} Tier-3 or Tier-4 is invoked. Suppose Tier-3 is a large model with specialized legal-medical training. On this second attempt, partial and final checks pass with $S_{\text{combined}}=0.9$.
        \item \textbf{Online Improvement:} The entire chain is logged, indicating Tier-2 legal-LLM needs improved medical references or a specialized legal-medical model. Over time, the system may refine thresholds or domain classification to skip Tier-2 for queries referencing both ISO-13485 and HIPAA.
      \end{enumerate}
  \end{itemize}
\end{itemize}

\end{document}